\documentclass[letterpaper,12pt,peerreviewca,draftcls]{IEEEtran}
\usepackage{csmMBFeb07,graphicx,url}

\usepackage{amssymb,amsmath,color,psfrag}
\usepackage{framed,graphicx,epsfig}
\usepackage{algorithm,algorithmicx,xspace}
\usepackage{indentfirst}
\usepackage{subfig,float}
\usepackage{multibib}
\usepackage{comment}
\usepackage{bm,times}
\usepackage{array}
\usepackage{multirow}

\usepackage{algpseudocode}

\usepackage{vmargin}
\setpapersize{USletter}
\setmarginsrb{1in}{1in}{1in}{1in}%
           {9pt}{9pt}{9pt}{18pt}

\newcommand\oprocendsymbol{\hbox{$\blacksquare$}}
\newcommand\oprocend{\relax\ifmmode\else\unskip\hfill\fi\oprocendsymbol}

\newcounter{sidebar}
\newcounter{sidebarequation}
\newcounter{sidebarproposition}
\newcounter{example}
\newcommand{\ouralg}{\textsl{Interim~Master~D-CL}\xspace}

\newcommand{\VV}{\mathcal{V}}

\newcommand{\real}{{\mathbb{R}}}
\newcommand{\reals}{{\mathbb{R}}}

\newcommand{\prpg}{\mbox{-}}
\newcommand{\updt}{\mbox{+}}

\newcommand{\until}[1]{\in\{1,\cdots,#1\}}

\newcommand{\vect}[1]{\boldsymbol{\mathbf{#1}}}
\newcommand{\Bvect}[1]{\bar{\boldsymbol{\mathbf{#1}}}}
\newcommand{\Tvect}[1]{\tilde{\boldsymbol{\mathbf{#1}}}}
\newcommand{\Hvect}[1]{\hat{\boldsymbol{\mathbf{#1}}}}

\newcommand{\Diag}[1]{\operatorname{Diag}(#1)}

\usepackage{tcolorbox}
\definecolor{mycolor}{rgb}{0.122, 0.435, 0.698}
\makeatletter
\newcommand{\mybox}[1]{%
  \setbox0=\hbox{#1}%
  \setlength{\@tempdima}{\dimexpr\wd0+13pt}%
  \begin{tcolorbox}[colframe=mycolor,boxrule=0.5pt,arc=4pt,
      left=6pt,right=6pt,top=6pt,bottom=6pt,boxsep=0pt,width=\@tempdima]
    #1
  \end{tcolorbox}
}
\makeatother

\begin{document}

\title{Cooperative Localization for Mobile Agents \\
{\Large A recursive
  decentralized algorithm based on\\ \vspace{-0.3in}Kalman filter decoupling}}
  
\author{Solmaz S. Kia, Stephen Rounds and Sonia Mart{\'\i}nez}

\maketitle

\CSMsetup

\begin{figure}[t]
 \begin{center}
 \includegraphics[trim=0 0 0 0pt,clip,height=2.6  in]{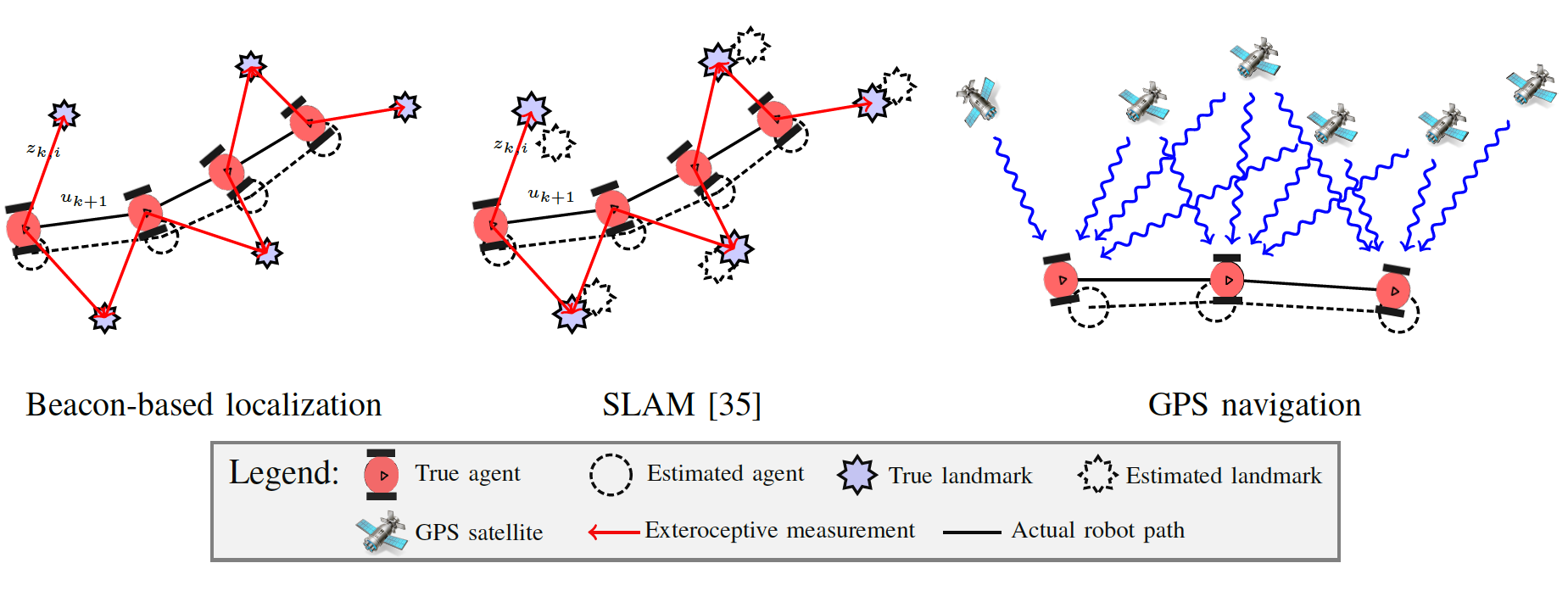}
 \end{center}
 \caption{{\small Schematic representation of common probabilistic localization
     techniques for mobile platforms: In beacon-based localization, the
     map of the area is known and there are pre-installed beacons or
     landmarks with known locations and identities. By taking
     relative measurements with respect to these landmarks, the mobile
     agents can improve their localization accuracy. For operations
     where a priori knowledge about the environment is not available,
     but nevertheless, the environment contains fixed and
     distinguishable features that agents can measure, SLAM is
     normally used to localize the mobile agents. SLAM is a process by
     which a mobile agent can build a map of an environment and at the
     same time use this map to deduce its location. On the other hand,
     GPS navigation provides location and time information in all
     weather conditions, anywhere on or near the earth but it requires
     an clear line of sight to at least four GPS
     satellites.}}\label{fig:local_tech}
\end{figure}

Technological advances in ad-hoc networking and miniaturization of
electro-mechanical systems are making possible the use of large
numbers of mobile agents (e.g., mobile robots, human agents, unmanned
vehicles) to perform surveillance, search and rescue, transport and
delivery tasks in aerial, underwater, space, and land environments.
However, the successful execution of such tasks often hinges upon
accurate position information, which is needed in lower level
locomotion and path planning algorithms. Common techniques for
localization of mobile robots are the classical pre-installed
beacon-based localization algorithms~\cite{JL-HFD:91}, fixed
feature-based Simultaneous Localization and Mapping (SLAM)
algorithms~\cite{MWMGD-PN-SC-HFDW-MC:01}, and GPS
navigation~\cite{SC-HDW:94}, see Fig.~\ref{fig:local_tech} for further
details.  However, in some operations such as search and
rescue~\cite{JSJ-GW-WFE:97,AD:02}, environment
monitoring~\cite{NB-JH-DE:00,MT-MR-SC-AL-HI-AJL:08}, and oceanic
exploration~\cite{AB-MRW-JJL:09}, the assumptions required by the
aforementioned localization techniques include the existence of
distinct and static features that can be revisited often, or clear
line-of-sight to GPS satellites. Such conditions may not be realizable
in practice, and thus these localization techniques become unfeasible.
Instead, Cooperative localization (CL) is emerging as an alternative
localization technique that can be employed in such scenarios.


In CL, a group of mobile agents with processing and communication
capabilities use relative measurements with respect to each other (no
reliance on external features) as a feedback signal to \emph{jointly
  estimate} the poses of all team members, which results in an
increased accuracy for the entire team. The particular appeal of CL
relies on the fact that  sporadic access to accurate localization
information by a particular robot results into a net benefit for the
rest of the team. This is possible thanks to the coupling that is
created through the state estimation process. Another nice feature of
CL is its cost effectiveness, as it does not require extra hardware
beyond the operational components normally used in cooperative robotic
tasks.  In such situations, agents are normally equipped with unique
identifiers and sensors which enable them to identify and locate other
group members. To achieve coordination, these agents often broadcast
their status information to one another.  In addition, given the wide
and affordable availability of communication devices, CL has also
emerged as an augmentation system to compensate for poor odometric
measurements, noisy and distorted measurements from other sensor
suites such as onboard IMU systems, see e.g.,~\cite{HM-DGE:14}.

The idea of exploiting relative robot-to-robot measurements for
localization can be traced back to~\cite{RK-SN-SH:94}, where members of
a mobile robotic team were divided into two groups, which took turns
remaining stationary as landmarks for the others. In later
developments in~\cite{IR-GD-EM:00}, where the term cooperative
localization was also introduced, the necessity for some robots to be
stationary was removed. Since then, many cooperative localization
algorithms using various estimation strategies such as Extended Kalman
filters (EKF)~\cite{SIR:00}, maximum likelihood~\cite{AH-MM-GS:02},
maximum a posteriori (MAP)~\cite{EDN-SIR-AM:09}, and particle
filters~\cite{DF-WB-HK-ST:00,AH-MJM-GSS:03,AP-AM:11} have been
developed. Cooperative localization techniques to handle system and measurement
models with non-Gaussian noises are also discussed
in~\cite{ATI-JWF-RLM-ASW:05,JN-DZ-IS-PH:13}.

Although CL is a very attractive concept for multi-robot
  localization, which does not require environmental features or GPS
  information, it also poses new challenges associated with the
  implementation of such a policy with acceptable communication,
  memory, and processing costs. Cooperative
localization is a joint estimation process which results in highly
coupled pose estimation for the full robotic team.  These
couplings/cross-correlations are created due to the relative
measurement updates. Accounting for these coupling/cross-correlations
is crucial for both filter consistency and also for propagating the
benefit of a robot-to-robot measurement update to the entire group.
In Section ``Cooperative localization via EKF'' we
  demonstrate these features in detail both through technical and
  simulation demonstrations.

A centralized implementation of CL is the most straightforward
mechanism to keep an accurate account of these couplings and, as a
result, obtain more accurate solutions. In a centralized scheme,
\emph{at every time-step}, a single device, either a leader robot or a
fusion center (FC), gathers and processes information from the entire
team. Then, it broadcasts back the estimated location results to each
robot (see e.g.,~\cite{SIR:00,AH-MJM-GSS:02}). Such a central
operation incurs into a high processing cost on the FC and a high
communication cost on both FC and each robotic team member. Moreover,
it lacks robustness that can be induced by single point failures. This
lack of robustness and energy inefficiency make the centralized
implementation less preferable.

A major challenge in developing decentralized CL (D-CL) algorithms is
how to maintain a precise account of cross-correlations and couplings
between the agents' estimates without invoking \emph{all-to-all}
communication \emph{at each time-step}.  The design and analysis of
decentralized CL algorithms, which maintain the consistency of the
estimation process while maintaining ``reasonable'' communication and
computation costs have been the subject of extensive research since
the CL idea's conception.  In Section ``Decentralized cooperative
localization: how to account for intrinsic correlations in cooperative
localization,'' we provide an overview of some of the D-CL algorithms
in the literature, with a special focus on how these algorithms
maintain/account for intrinsic correlations of CL
strategy. We provide readers a more technical example of
  a D-CL algorithm in the Section ``The \ouralg algorithm: a tightly
  coupled D-CL strategy based on Kalman filter decoupling,'' which is
  a concise summary of the solution in~\cite{SSK-SR-SM:14-iros}
  developed by the authors.

The reader interested on technical analysis and details beyond
decentralization for CL can find a brief literature guide in ``Further
Reading.''

\noindent\textbf{Notations}: Before proceeding further, let us introduce our notations.
We denote by $\mathbb{M}_n$, $\vect{0}_{n\times m}$ (when $m=1$, we
use $\vect{0}_n$) and $\vect{I}_n$, respectively, the set of real
positive definite matrices of dimension $n\times n$, the zero matrix
of dimension $n\times m$, and the identity matrix of dimension
$n\times n$. We represent the transpose of matrix
$\vect{A}\in\reals^{n\times m}$ by $\vect{A}^\top$.  The block
diagonal matrix of set of matrices $\vect{A}_1,\dots,\vect{A}_N$ is
$\mathrm{Diag}(\vect{A}_1,\cdots,\vect{A}_N)$. For finite sets $V_1$
and $V_2$, $V_1\backslash V_2$ is the set of elements in $V_1$, but
not in $V_2$. For a finite set $V$ we represent its cardinality by
$|V|$. In a team of $N$ agents, the local variables associated with
agent $i$ are distinguished by the superscript $i$, e.g., $\vect{x}^i$
is the state of agent $i$, $\Hvect{x}^i$ is its state estimate, and
$\vect{P}^{i}$ is the covariance matrix of its state estimate. We use
the term \emph{cross-covariance} to refer to the correlation terms
between two agents in the covariance matrix of the entire network. The
cross-covariance of the state vectors of agents $i$ and $j$ is
$\vect{P}_{ij}$. We denote the propagated and updated variables, say
$\Hvect{x}^i$, at time-step $k$ by $\Hvect{x}^{i\prpg}(k)$ and
$\Hvect{x}^{i\updt}(k)$, respectively.  We drop the time-step argument
of the variables as well as matrix dimensions whenever they are clear
from the context.  In a network of $N$ agents, $\vect{p} =
(\vect{p}^1,\dots,\vect{p}^N) \in \real^d$, $d={\sum_{i=1}^Nn^i}$ is
the aggregated vector of local vectors $\vect{p}^i\in\real^{n^i}$.

\section{Cooperative localization via EKF}
This section provides an overview of a CL strategy that employs an EKF
following~\cite{SIR-GAB:02}. By a close examination of this algorithm,
it is possible to explain why accounting for the intrinsic
cross-correlations in CL is both crucial for filter consistency and
key to transmit the benefit of an update of a relative robot-to-robot
measurement to the entire team. We also discuss the computational cost
of implementing this algorithm in a centralized manner.

\begin{figure}[t]
\centering
 \includegraphics[trim=0 0 0 0pt,clip,scale=0.27]{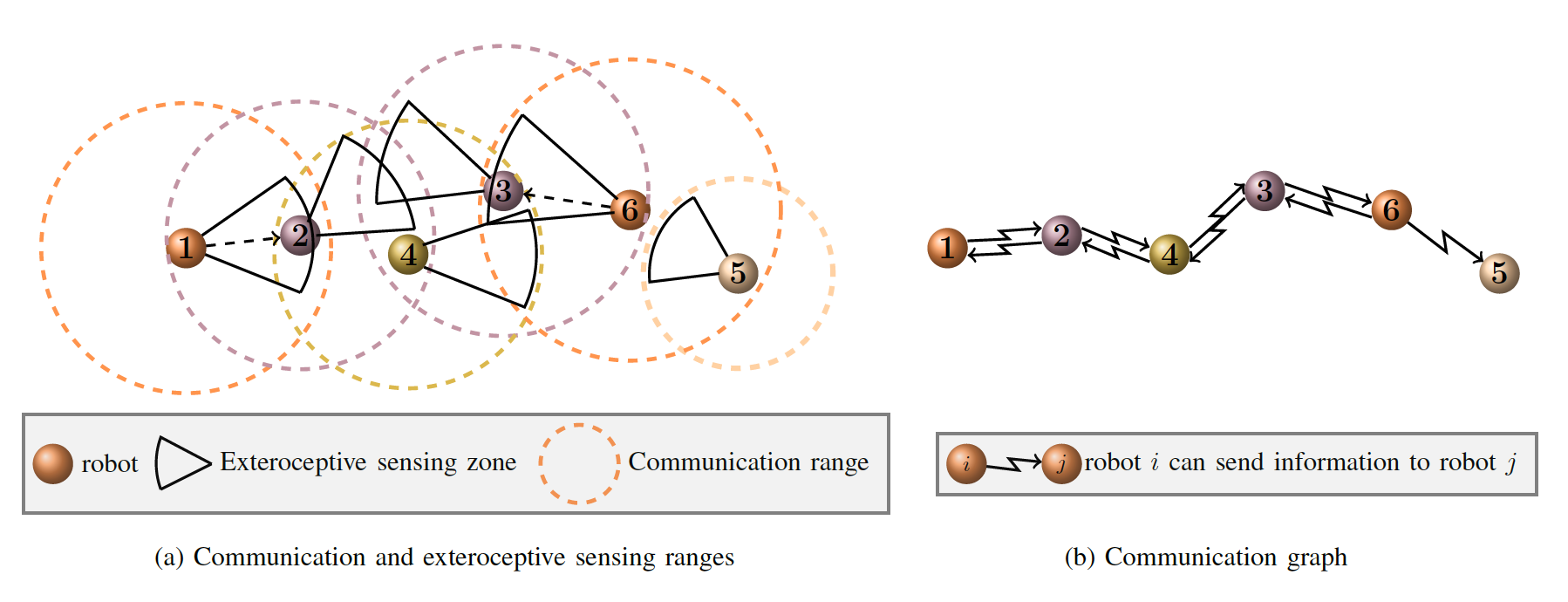}
\caption{{\small This figure depicts a multi-hop communication
    scenario for the multi-robot team. Plot (a) shows the
    communication and measurement ranges. Here, robots $1$ and $6$
    make relative measurements, respectively, of robots $2$ and
    $3$. Plot (b) shows the communication graph generated using the
    communication ranges given in plot (a). Here the robot at the head
    of an arrow can send information to the robot at the tip of the
    arrow.  As this graph shows, each of robots $1$ and $6$ can pass
    communication message to the entire team via a multi-hop strategy.
  }}\label{fig::data_propagation} \vspace{-0.2in}
\end{figure}

First, we briefly describe the model considered for the mobile robots
in the team.  Consider a group of $N$ mobile agents with
communication, processing and measurement capabilities.  Depending on
the adopted CL algorithm, communication can be in bidirectional manner
with a fusion center, a single broadcast to the entire team or in
multi-hop fashion as shown in Fig.~\ref{fig::data_propagation}, i.e.,
every agent re-broadcasts every received message intended to reach the
entire team. Each agent has a detectable unique identifier (UID)
which, without loss of generality, we assume to be a unique integer
belonging to the set $\VV=\{1,\dots,N\}$.  Using a set of so-called
proprioceptive sensors every agent $i\in\VV$ measures its self-motion,
for example by compass readings and/or wheel encoders, and uses it to
propagate its equations of motion.

\begin{align}\label{eq::RobotMotionModelNoise-indv}
  \vect{x}^i(k+1)&= \vect{f}^i(\vect{x}^i(k),
  \vect{u}^i(k))+\vect{g}^i(\vect{x}^i(k))\vect{\eta}^i(k),
\end{align}
where $\vect{x}^i\in\reals^{n^{i}}$, $\vect{u}^i\in\reals^{m^i}$, and
$\vect{\eta}^i\in\reals^{{p^i}}$ are, respectively, the state vector,
the input vector and the process noise vector of agent $i$. Here,
$\vect{f}^i(\vect{x}^i,\vect{u}^i)$ and $\vect{g}^i(\vect{x}^i)$, are,
respectively, the system function and process noise coefficient
function of the agent $i\in\VV$. The state vector of each agent can be
composed of variables that describe the robots global pose in the
world (e.g. latitude, longitude, direction), as well as other
variables potentially needed to model the robots dynamics
(e.g. steering angle, actuation dynamics).  The team can consist of
heterogeneous agents, nevertheless, the collective motion equation of
the team can be represented by
\begin{align}\label{eq::RobotMotionModelNoise}
\vect{x}(k+1)&=
\vect{f}(\vect{x}(k),\vect{u}(k))+\vect{g}(\vect{x}(k))\vect{\eta}(k),
\end{align}
where, $\vect{x}=(\vect{x}^1,\cdots,\vect{x}^N)$,
$\vect{u}=(\vect{u}^1,\cdots,\vect{u}^N)$,
$\vect{\eta}=(\vect{\eta}^1,\cdots,\vect{\eta}^N)$,
$\vect{f}(\vect{x},\vect{u})=(\vect{f}^1(\vect{x}^1,\vect{u}^1),
\cdots,\vect{f}^N(\vect{x}^N,\vect{u}^N))$ and
$\vect{g}(\vect{x})=\Diag{\vect{g}^1(\vect{x}^1),\cdots,
  \vect{g}^N(\vect{x}^N)}$.

Obviously, if each agent only relies on propagating its equation of
motion in~\eqref{eq::RobotMotionModelNoise-indv} using self-motion
measurements, this state estimate  grows unbounded due to the noise
term $\vect{\eta}^i(k)$. To reduce the growth rate of this estimation
error, a CL strategy can be employed.  Thus, let every agent $i\in\VV$
also carry exteroceptive sensors to monitor the environment to detect,
uniquely, the other agents $j\in\VV$ in the team and take relative
measurements
\begin{align}\label{eq::measur_ij}
  \vect{z}_{ij}(k+1)& =\vect{h}_{ij}(\vect{x}^i(k),\vect{x}^j(k))+
  \vect{\nu}^i(k),
\end{align}
where $\vect{z}_{ij}\in\real^{n_z^i}$ from them, e.g., relative pose,
relative range, relative bearing measurements, or both. Here,
$\vect{h}_{ij}(\vect{x}^i,\vect{x}^j)$ is the measurement model and
$\vect{\nu}^i$ is the measurement noise of agent $i\in\VV$.
Relative-measurement feedback, as shown below, can help the agents
improve their localization accuracy, though the overall uncertainty
can not be bounded (c.f.~\cite{SIR-GAB:02}).  The tracking performance
can be improved significantly if agents have occasional absolute
positioning information, e.g., via GPS or relative measurements taken
from a fixed landmark with a priori known absolute location. Any
absolute pose measurement by an agent $i\in\VV$, e.g., through
intermittent GPS access, is modeled by $ \vect{z}_{ii}(k+1)
=\vect{h}_{ii}(\vect{x}^i(k))+ \Bvect{\nu}^i(k)$.  The agents can
obtain concurrent exteroceptive absolute and relative measurements.

Let us assume all the process noises $\vect{\eta}^i$ and the
measurement noise $\vect{\nu}^i$, $i\in\VV$, are independent zero-mean
white Gaussian processes with, respectively, known positive definite
variances $\vect{Q}^i(k)=E[ \vect{\eta}^i(k){\vect{\eta}^i}(k)^\top]$,
$\vect{R}^i(k)=E[ \vect{\nu}^i(k){\vect{\nu}^i}(k)^\top]$ and
$\Bvect{R}^i(k)=E[ \Bvect{\nu}^i(k){\Bvect{\nu}^i}(k)^\top]$.
Moreover, let all the sensor noises be white and mutually uncorrelated
and all sensor measurements be synchronized.  Then, the centralized
EKF CL algorithm is a straightforward application of EKF over the
collective motion model of the robotic
team~\eqref{eq::RobotMotionModelNoise} and measurement
model~\eqref{eq::measur_ij}.  The propagation stage of this algorithm
is
 \begin{subequations}\label{eq::propag_central}
 \begin{align}
   \Hvect{x}^{\prpg}(k+1)&=
   \vect{f}(\Hvect{x}^{\updt}(k),\vect{u}(k)),\label{eq::propag_central-a}\\
   \vect{P}^{\prpg}(k+1)&=
   \vect{F}(k)\vect{P}^{\updt}(k)\vect{F}(k)^\top+\vect{G}(k)
   \vect{Q}(k)\vect{G}(k)^\top.\label{eq::propag_central-b}
\end{align}
 \end{subequations}
 where $\vect{F}=\Diag{\vect{F}^1,\cdots,\vect{F}^N}$,
 $\vect{G}=\Diag{\vect{G}^1,\cdots,\vect{G}^N}$ and
 $\vect{Q}=\Diag{\vect{Q}^1,\cdots,\vect{Q}^N}$, with, for all
 $i\in\VV$, $\vect{F}^i=\frac{\partial}{\partial
   \vect{x}^i}\vect{f}(\Hvect{x}^{i\updt}(k),\vect{u}^i(k))$ and
 $\vect{G}^i=\frac{\partial}{\partial
   \vect{x}^i}\vect{g}(\Hvect{x}^{i\updt}(k))$.
  
 If there exists a relative measurement in the network at some given
 time $k+1$, say robot $a$ takes relative measurement from robot $b$, the states are updated as follows. The innovation of the
 relative
 measurement  and its covariance are,
 respectively, 
\begin{subequations}\label{eq::EKF-measur-eq}
\begin{align}
  \vect{r}^{a}&=\vect{z}_{ab}- \vect{h}_{ab}(\Hvect{x}^{a\prpg}(k+1),
  \Hvect{x}^{b\prpg}(k+1)),\label{eq::reletive_Residual}\\
  \!\!\vect{S}_{ab}&\!=
  \!\vect{H}_{ab}(k\!+\!1)\vect{P}^{\prpg}(k\!+\!1)
  \vect{H}_{ab}(k\!+\!1)^\top\!\!+\!\vect{R}^a(k\!+\!1).\label{eq::S_abmain}
\end{align}
\end{subequations} 
where (without loss of generality we let $a<b$)
\begin{align}
  &\vect{H}_{ab}(k)=\big[\overset{1}
  {\vect{0}}~~\overset{\cdots}{\cdots}~~\overset{a}{-\Tvect{H}_a}(k)~~\overset{a+1}{\vect{0}}~~\overset{\cdots}{\cdots}~~\overset{b}{\Tvect{H}_b}(k)~~\overset{b+1}{\vect{0}}~~\overset{\cdots}{\cdots}\big],\nonumber\\
  &\Tvect{H}_a(k)=-\frac{\partial}{\partial \vect{x}^a}\vect{h}_{ab}(\Hvect{x}^{a\prpg}(k),\Hvect{x}^{b\prpg}(k)),\label{eq::H_ab} \\
  &\Tvect{H}_b(k)=\frac{\partial}{\partial
    \vect{x}^b}\vect{h}_{ab}(\Hvect{x}^{a\prpg}(k),\Hvect{x}^{b\prpg}(k)).\nonumber
\end{align} 

An absolute measurement by a robot $a\in\VV$ can be processed
similarly, except that in~\eqref{eq::H_ab}, $\Tvect{H}_b$ becomes
zero, while in~\eqref{eq::EKF-measur-eq}, the index $b$ should be
replaced by $a$ and $\vect{R}^a(k\!+\!1)$ should be replaced by
$\Bvect{R}^a(k\!+\!1)$.  Then, the Kalman filter gain is given
by
\begin{equation*}
  \vect{K}(k+1)=\vect{P}^{\prpg}(k+1)\vect{H}_{ab}(k+1)^\top
  {\vect{S}_{ab}}^{-1}.
\end{equation*} 
And, finally, the collective pose update and covariance update
equations for the network are:
\begin{subequations}\label{eq::update_central}
\begin{align}
  \Hvect{x}^{\updt}(k\!+\!1)=&\Hvect{x}^{\prpg}(k\!+\!1)\!+\!\vect{K}(k\!+\!1)\vect{r}^{a},\label{eq::update_central-a}\\
  \vect{P}^{\updt}(k\!+\!1)=&\vect{P}^{\prpg}(k\!+\!1)\!-\!\vect{K}(k\!+\!1)\vect{S}_{ab}\vect{K}(k\!+\!1)^\top.\!\label{eq::update_central-b}
\end{align}
\end{subequations} 
Because $\vect{K}(k\!+\!1)\vect{S}_{ab}\vect{K}(k\!+\!1)^\top$ is a
positive semi-definite term, the update
equation~\eqref{eq::update_central-b} clearly shows that any relative
measurement update results in a reduction of the estimation
uncertainty.

To explore the relationship among the estimation equations of each robot,
we express the aforementioned collective form of the EKF CL in terms
of its agent-wise components, as shown in
Algorithm~\ref{alg::CEKFCL}. Here, the Kalman filter gain is
partitioned into
$\vect{K}=\begin{bmatrix}\vect{K}_1^\top,\cdots,\vect{K}_N^\top
\end{bmatrix}^\top$, where $\vect{K}_i\in\real^{n^i\times n_z^i}$ is
the portion of the Kalman gain used to update the pose estimate of the
agent $i\in\VV$.  To process multiple synchronized measurements,
\emph{sequential updating} (c.f.~for example
\cite[Ch.~3]{CTL:66},\cite{YB-PKW-XT:11}) is employed.


Algorithm~\ref{alg::CEKFCL} clearly showcases the role of past
correlations in a CL strategy.  First, observe that, despite having
decoupled equations of motion, the source of the coupling in the
propagation phase is the cross-covariance
equation~\eqref{eq::propag_central_Expanded-c}. Upon an incidence of a
relative measurement between agents $a$ and $b$, this term becomes
non-zero and its evolution in time requires the information of these
two agents. Thus, these two agents have to either communicate with
each other all the time or a central operator has to take over the
propagation stage. As the incidences of relative measurements grow,
more non-zero cross-covariance terms are created. As a result, the
communication cost to perform the propagation grows, requiring the
data exchange all the time with either a Fusion Center (FC) or
all-to-all agent communications, even when there is no relative
measurement in the network. The update
equations~\eqref{eq::RobotCovarUpdate} are also coupled and their
calculations need, in principle, a FC. The next observation regarding
the role of the cross-covariance terms can be deduced from studying
the Kalman gain equation~\eqref{eq::KalmanKi}. As this equation shows,
when an agent $a$ takes a relative measurement from agent $b$, any
agent whose pose estimation is correlated with either of agents $a$
and $b$ in the past, (i.e., $\vect{P}_{i b}^{\prpg}(k+1)$ and/or
$\vect{P}_{ia}^{\prpg}(k+1)$ are non-zero) has a non-zero Kalman gain
and, as a result, the agent benefits from this measurement update. The
same is true in the case of an absolute measurement taken by a
robot~$a$.

\begin{figure}[htbp]
    \captionsetup[subfigure]{labelformat=empty}
      \unitlength=0.5in 
  \captionsetup[subfloat]{captionskip=5pt}
\begin{center}
  \subfloat[ ] {\includegraphics[trim=0 0 0 0pt,clip,height=1.5
    in,width=0.5\textwidth]{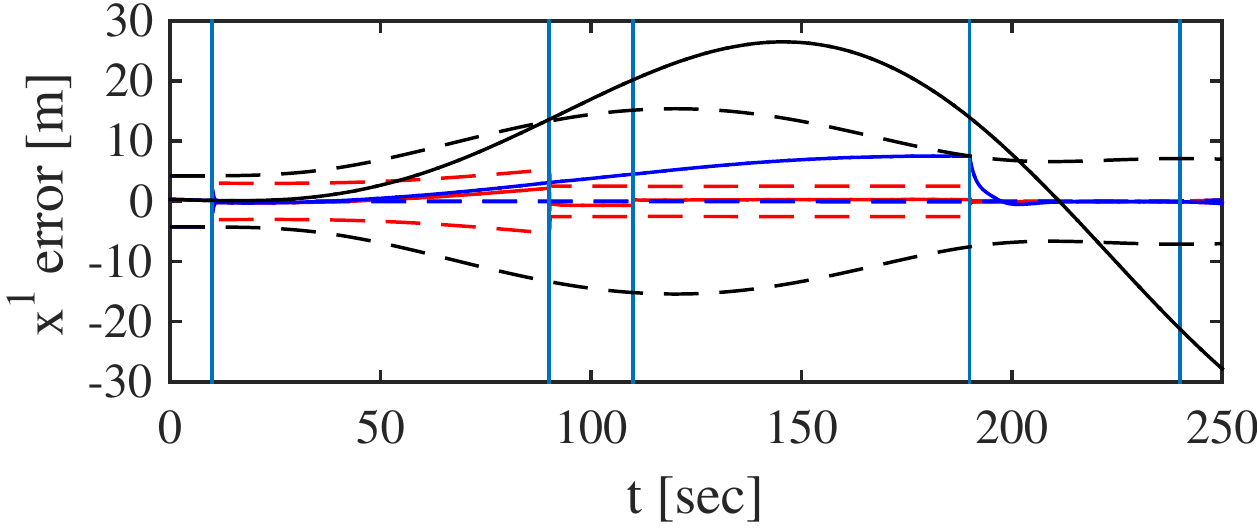}} \subfloat[ ]
  {\includegraphics[trim=0 0 0 0pt,clip,height=1.5
    in,width=0.5\textwidth]{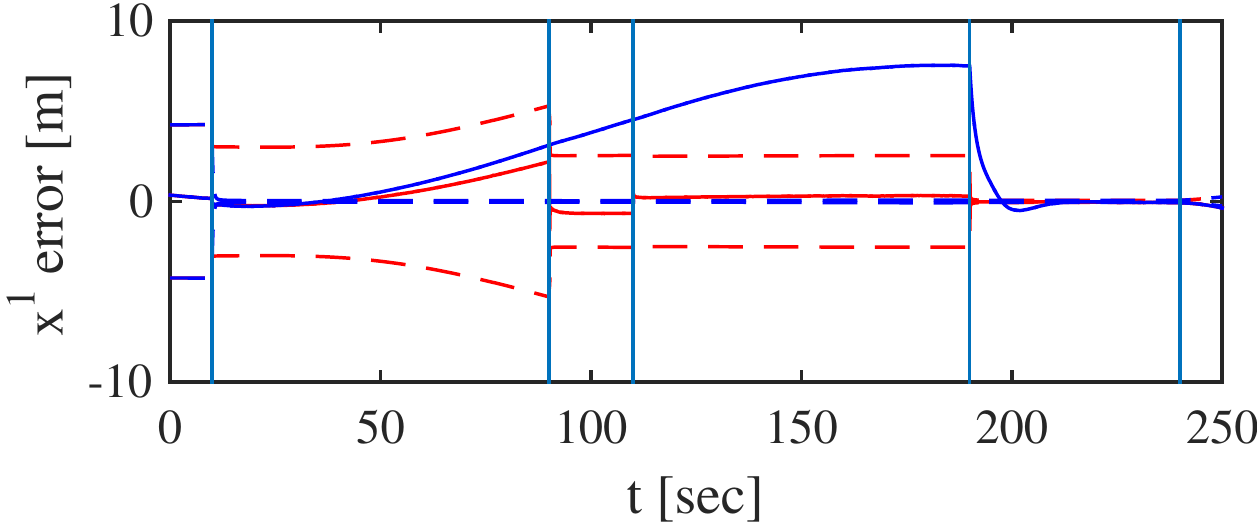}}
  \vspace{-2in} {\scriptsize$\quad \quad\quad ~~~~ \quad\quad \quad
    \quad\quad1\!\to\! 2\quad\quad~\,3\!\!\to\!\!1~\, 1\!\! \to\!\!3,
    3\!\!\to\! \!2,2\!\!\to\! \!1\quad\quad 1\!\to\!1$ $~~\quad
    \quad\quad ~~~~ \quad\quad \quad \quad\quad1\!\to\!
    2\quad\quad~\,3\!\!\to\!\!1~\, 1\!\! \to\!\!3, 3\!\!\to\!
    \!2,2\!\!\to\! \!1\quad\quad 1\!\to\!1$} \vspace{1.7in}\\\vspace{-0.1in}
   \subfloat[ ] {\includegraphics[trim=0 0 0
    0pt,clip,height=1.5
    in,width=0.5\textwidth]{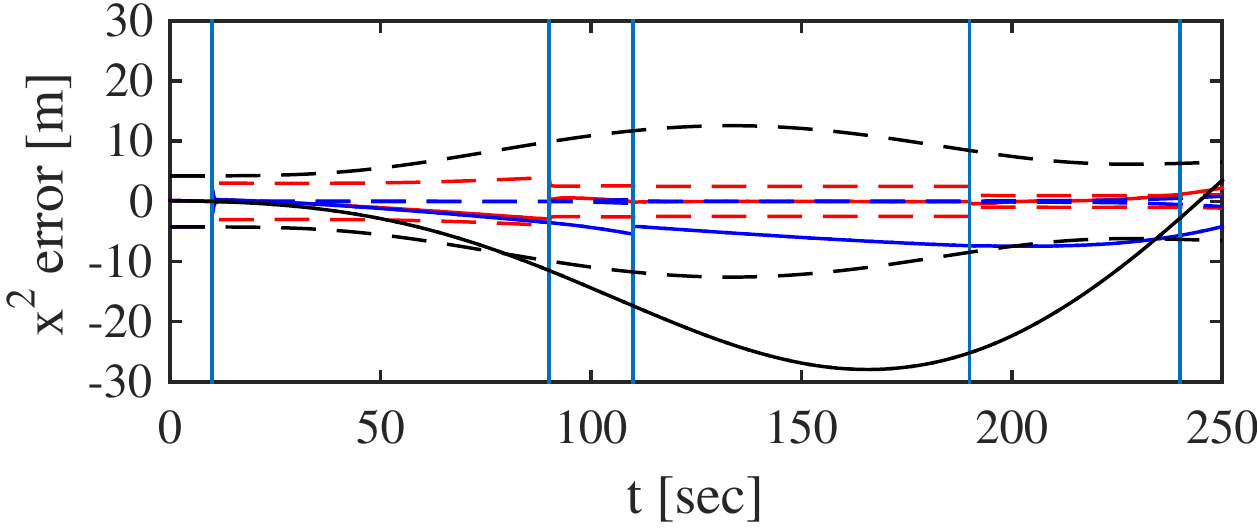}} \subfloat[ ]
  {\includegraphics[trim=0 0 0 0pt,clip,height=1.5
    in,width=0.5\textwidth]{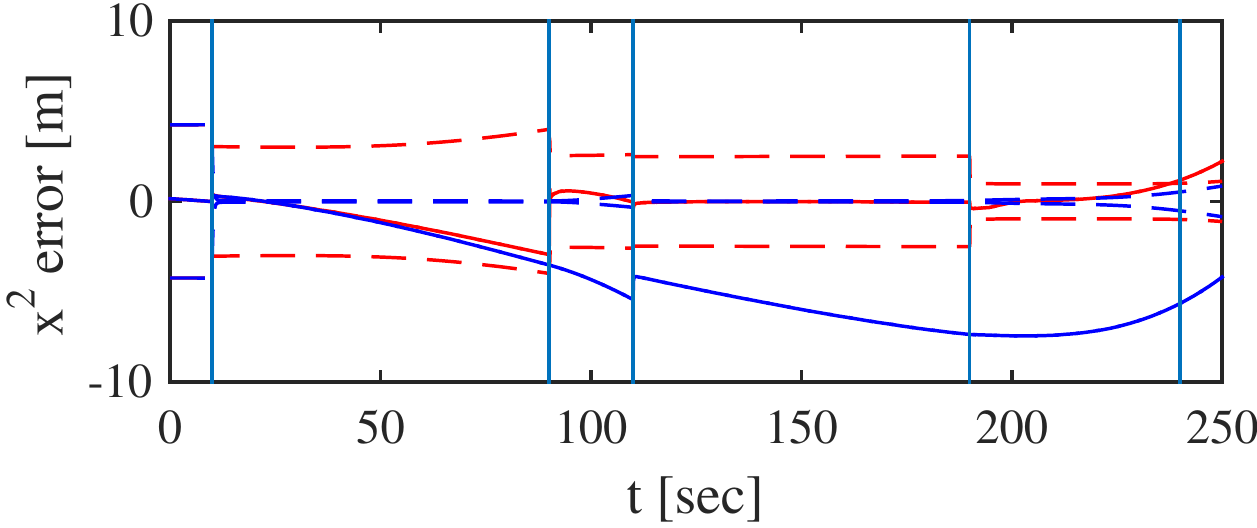}}\\\vspace{-0.1in}
  \subfloat[ ] {\includegraphics[trim=0 0 0
    0pt,clip,height=1.5
    in,width=0.5\textwidth]{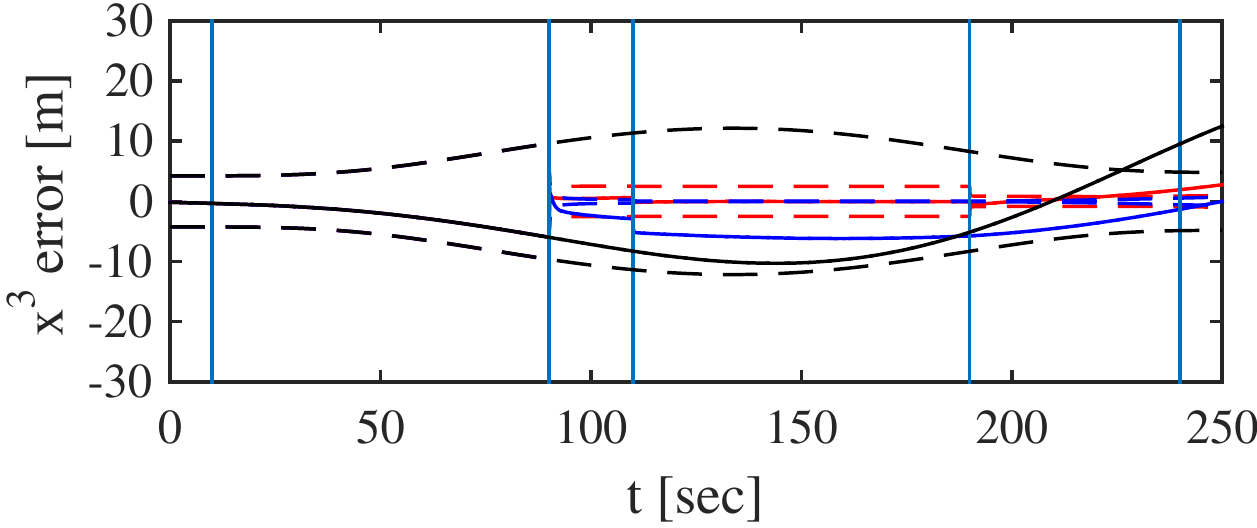}} \subfloat[ ]
  {\includegraphics[trim=0 0 0 0pt,clip,height=1.5
    in,width=0.5\textwidth]{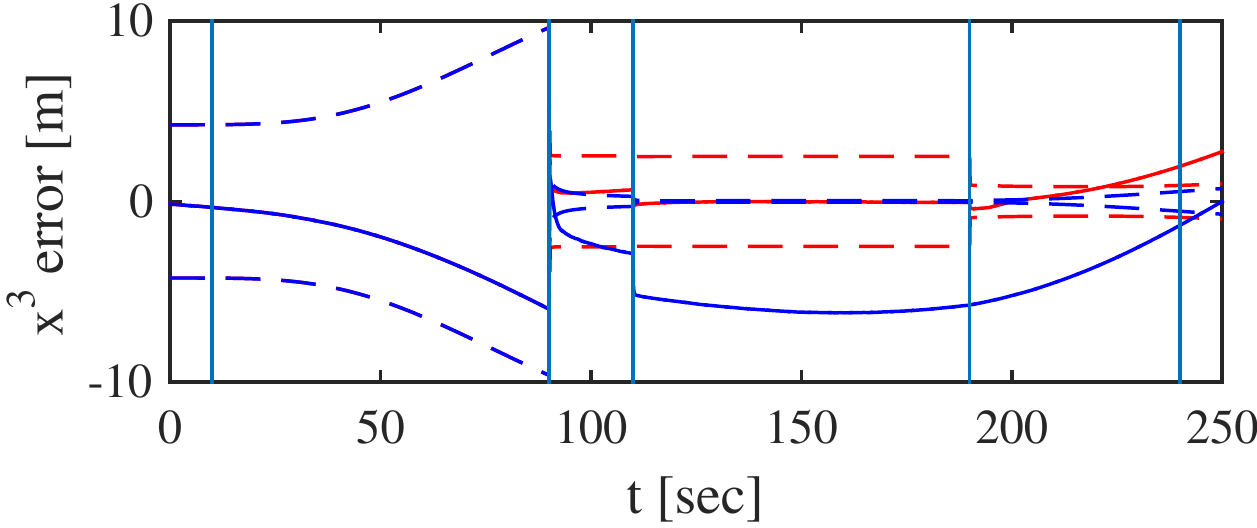}}
   \end{center}\vspace{-0.7in}        
   \caption{{\small Estimation error (solid line) and $3\sigma$ error
       bounds (dashed lines) in the $x-$coordinate variable for 3
       robots moving on a flat terrain when they (a) only propagate
       their equations of motion using self-motion measurements (black
       plots), (b) employ cooperative localization ignoring past
       correlations between the estimations of the robots (blue
       plots), (c) employ cooperative localization with accurate
       account of past correlations (red plots). The figures on the
       right column are the same figures as on the left where the
       localization case (a) is removed for clearer demonstration of
       cases (b) and (c). Here, $a\to b$ over the time interval marked
       by two vertical blue lines indicates that robot $a$ has taken a
       relative measurement with respect to robot $b$ at that time
       interval. The symbol $a\to a$ means that robot $a$ obtains an
       absolute measurement.}}\label{eq::fig_example}
          \end{figure}

The following simple simulation study demonstrates the significance of
maintaining an accurate account of cross-covariance terms between the
state estimates of the team members. We consider a team of $3$ mobile
robots moving on a flat terrain whose equations of motion in a fixed
reference frame, for $i\in\{1,2,3\}$, are modeled as
\begin{align*}
  x^i(k+1)=&x^i(k)+V^i(k)\cos(\phi(k))\,\delta t,\\
  y^i(k+1)=&y^i(k)+V^i(k)\sin(\phi(k))\,\delta t,\\
  \phi^i(k+1)=&\phi^i(k)+\omega(k)\,\delta t,
\end{align*}
where $V^i(k)$ and $\omega^i(k)$ are true linear and rotational velocities
of robot $i$ at time $k$ and $\delta t$ is the stepsize. Here, the
pose vector of each robot is
$\vect{x}^i=[x^i,\,y^i,\,\phi^i]^\top$. Every robot uses odometric
sensors to measure its linear $V^i_m(k)=V^i(k)+\eta^i_V(k)$ and
rotational $\omega^i_m(k)=\omega^i(k)+\eta^i_\omega(k)$, velocities,
where $\eta_V^i$ and $\eta_\omega^i$ are their respective
contaminating measurement noise. The standard deviation of
$\eta_V^i(k)$, $i\in\{1,2,3\}$, is $0.1V^i(k)$, while the standard
deviation of $\eta_\omega^i$ is $1\,\text{deg}/s$, for robot $1$ and
robot $2$, and $0.5\,\text{deg}/s$ for robot $3$. Robots $\{1,2,3\}$
can take relative pose measurements from one another. Here, we use
standard deviations of, respectively $(0.05 \,\text{m},0.05\,\text{m},
1\,\text{deg}/s)$, $(0.05 \,\text{m},0.05\,\text{m},
2\,\text{deg}/s)$, $(0.07\, \text{m},0.07\,\text{m},
1.5\,\text{deg}/s)$ for measurement noises. Assume robot $1$ can
obtain absolute position measurement with a standard deviation of
$(0.1 \text{m},0.1\text{m})$ for the measurement noise.
Figure~\ref{eq::fig_example} demonstrates the $x$-coordinate
estimation error (solid line) and the $3\sigma$ error bound (dashed
lines) of these robots when they (a) only propagate their equations of
motion using self-motion measurements (black plots), (b) employ an EKF
CL ignoring past correlations between the estimations of the robots
(blue plots), (c) employ an EKF CL with an accurate account of past
correlations (red plots). As this figure shows, employing a CL
strategy improves the localization accuracy by reducing both the
estimation error and its uncertainty. However, as plots in blue show,
ignoring the past correlations (here cross-covariances) among the
robots state estimates results in overly optimistic estimations
(notice the almost vanished $3\sigma$ error bound in blue plots while
the solid blue line goes out of these bounds, an indication of
inconsistent estimation). In contrast, by taking into account the past
correlations (see red plots), one sees a more consistent
estimation.

\begin{figure}[htbp]
    \captionsetup[subfigure]{labelformat=empty}
      \unitlength=0.5in 
  \captionsetup[subfloat]{captionskip=5pt}
\begin{center}
  \subfloat[ {\small Robot 1}] {\includegraphics[trim=0 0 0 0pt,clip,height=1.6
    in]{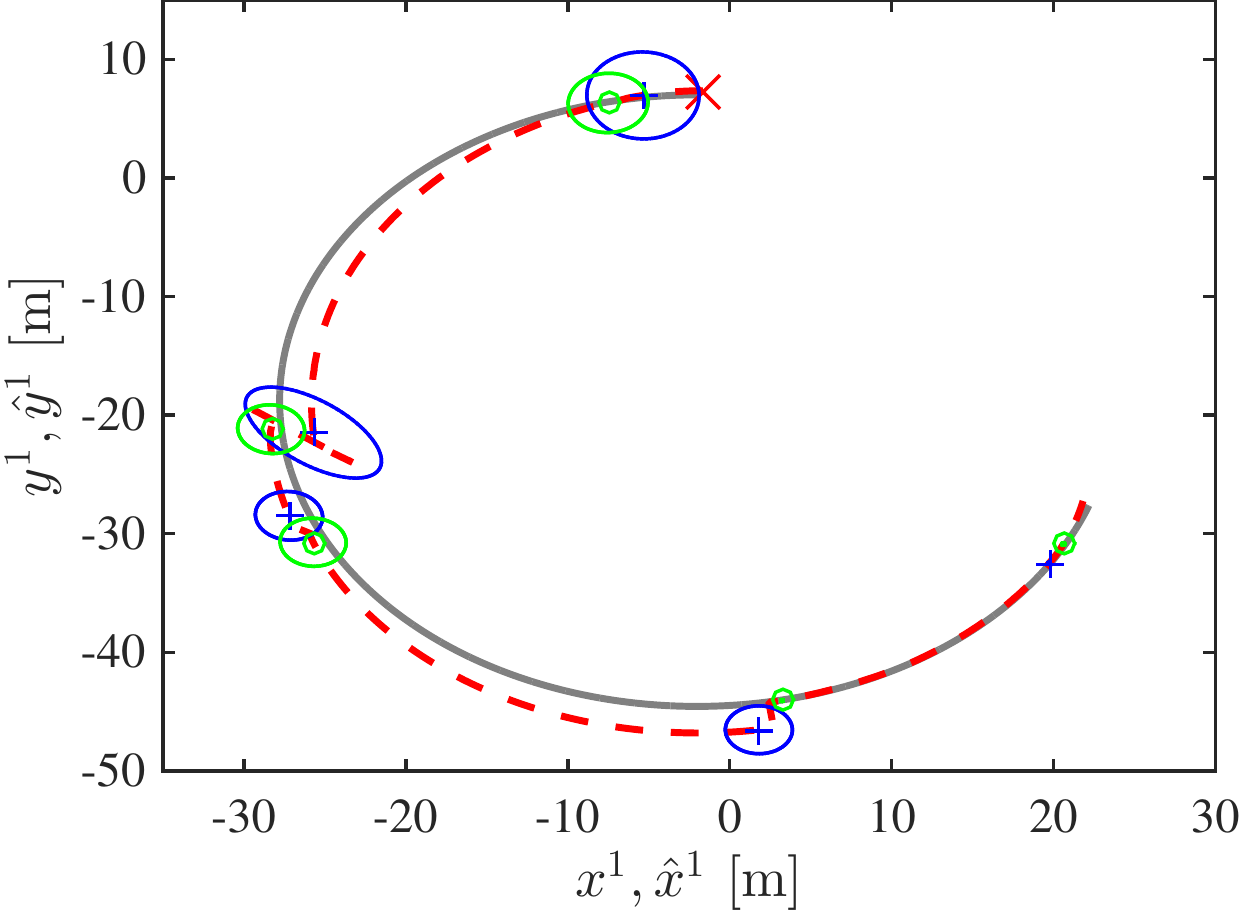}} \subfloat[{\small Robot 2}]
  {\includegraphics[trim=0 0 0 0pt,clip,height=1.6
    in]{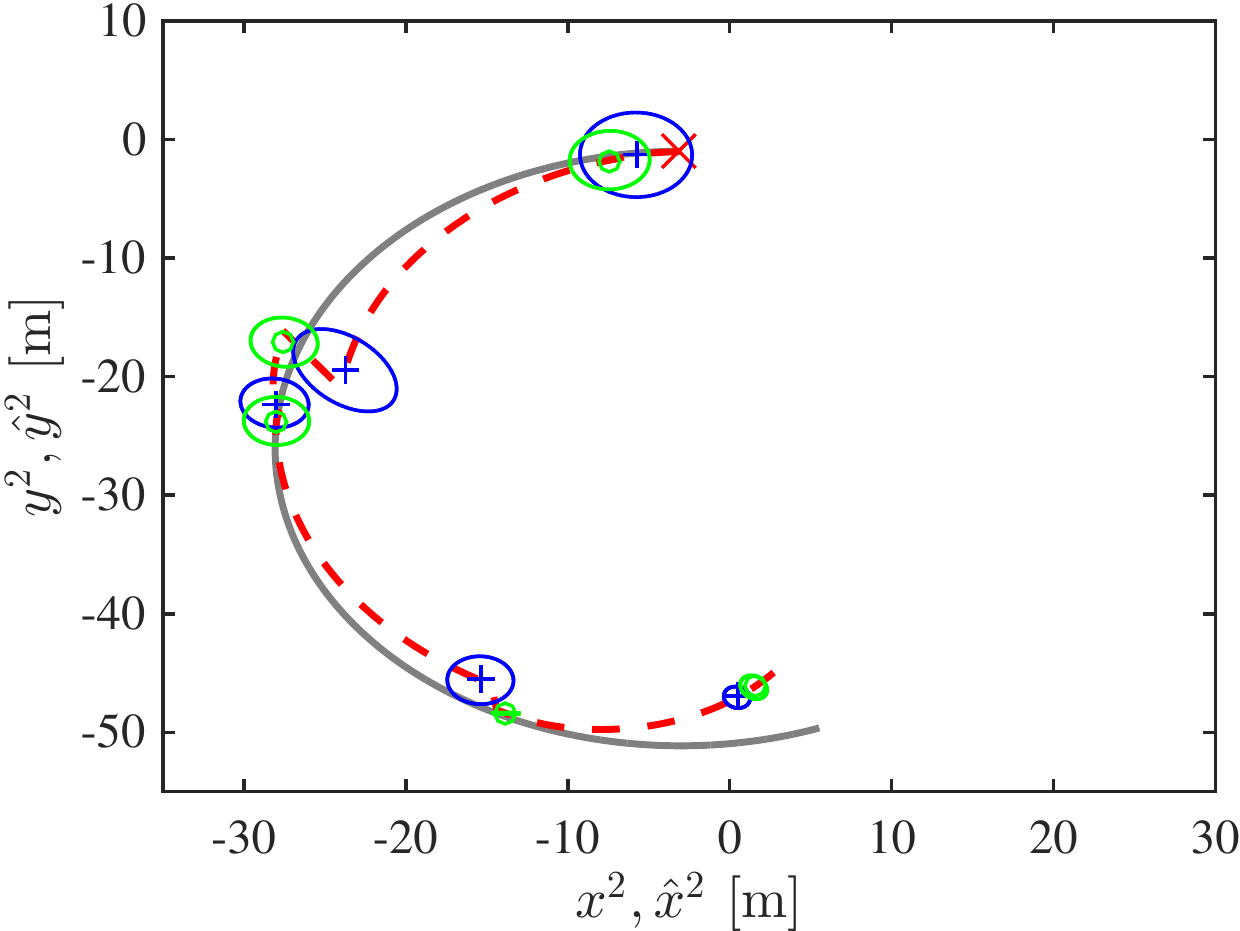}}
 \subfloat[ {\small Robot3}] {\includegraphics[trim=0 0 0
    0pt,clip,height=1.6
    in]{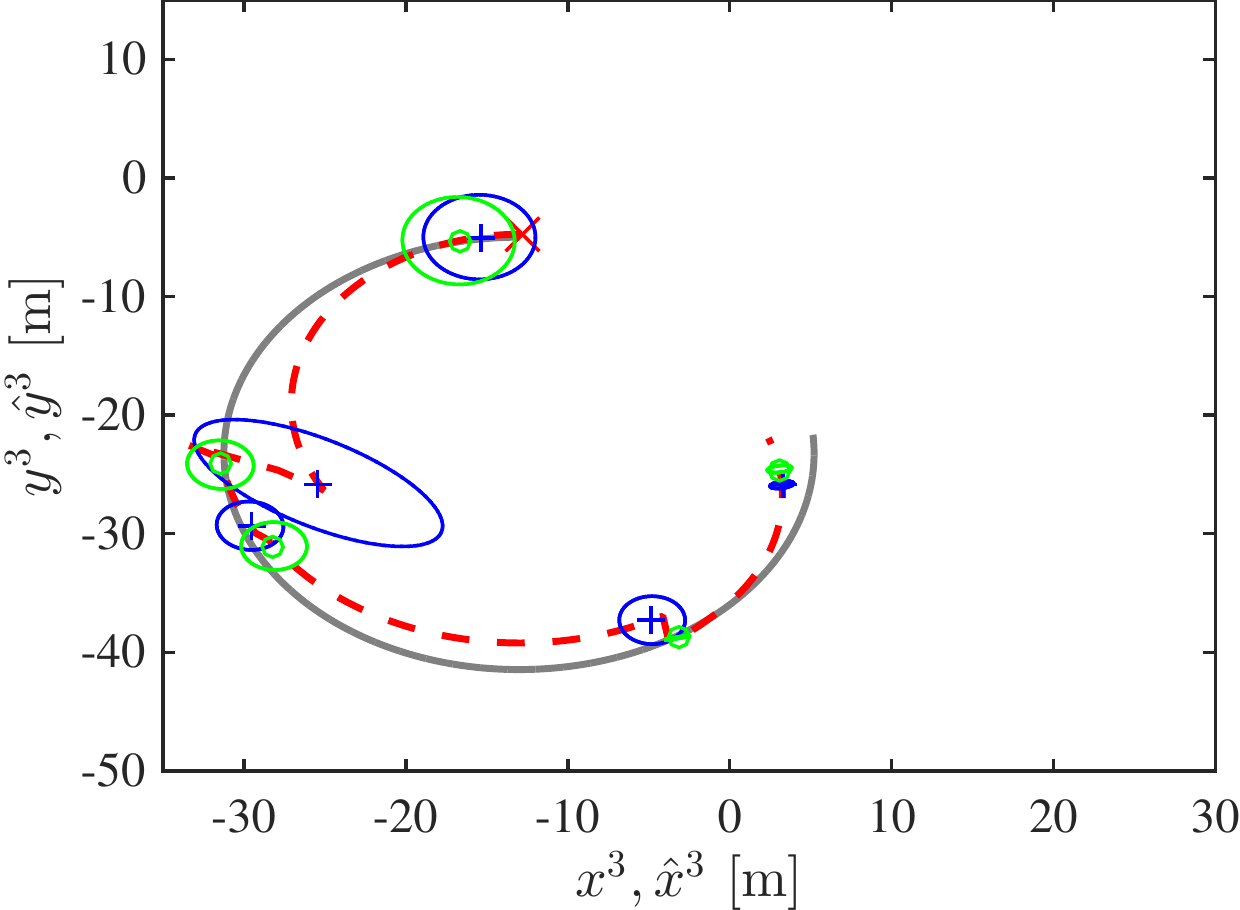}} 
   \end{center}
        
   \caption{{\small Trajectories of the robots for the simulation
       study of Fig.~\ref{eq::fig_example}. Here, the gray curve is
       the ground truth. The red curve is the estimation of the
       trajectory by implementing a EKF CL. The blue
       (resp. green) ellipses show the 95\% uncertainty regions for
       the estimations at 2 seconds before (resp. after) any change in
       the measurement scenario (see
       Fig.~\ref{eq::fig_example})}}\label{eq::fig_example2}
          \end{figure}

Figure~\ref{eq::fig_example}  also showcases the role of past cross-covariances to expand the
benefit of a relative measurement between two robots, or of an
absolute measurement by a robot to the rest of the team. For example
consider robot 2. In the time interval $[10,90]$ seconds, robot $1$ is
taking a relative measurement from robot $2$. As a result, the state
estimation equation of robot $1$ and robot $2$ are correlated, i.e,
the cross-covariance term between these two robots is
non-zero. Therefore, in the time interval $[90,110]$ seconds, when the
estimation update is due to the relative measurement taken by robot
$3$ from robot $1$, the estimation of robot $2$ is also improved (see
red plots.) In the time interval $[190,240]$ seconds, when the
estimation update is due to the absolute measurement taken by robot
$1$, robot $2$ and $3$ also benefit from this measurement update due
to past correlations (see the red plots.)
Figure~\ref{eq::fig_example2} shows the trajectories of the robots
when they apply EKF CL strategy. For more enlightening simulation
studies, we refer the interested reader to~\cite{SIR-GAB:02}.

\section{Decentralized cooperative localization: how to account for intrinsic correlations in cooperative localization}
\begin{figure}[htbp]
\begin{center}
\includegraphics[trim=0 0 0 0pt,clip,height=3
    in]{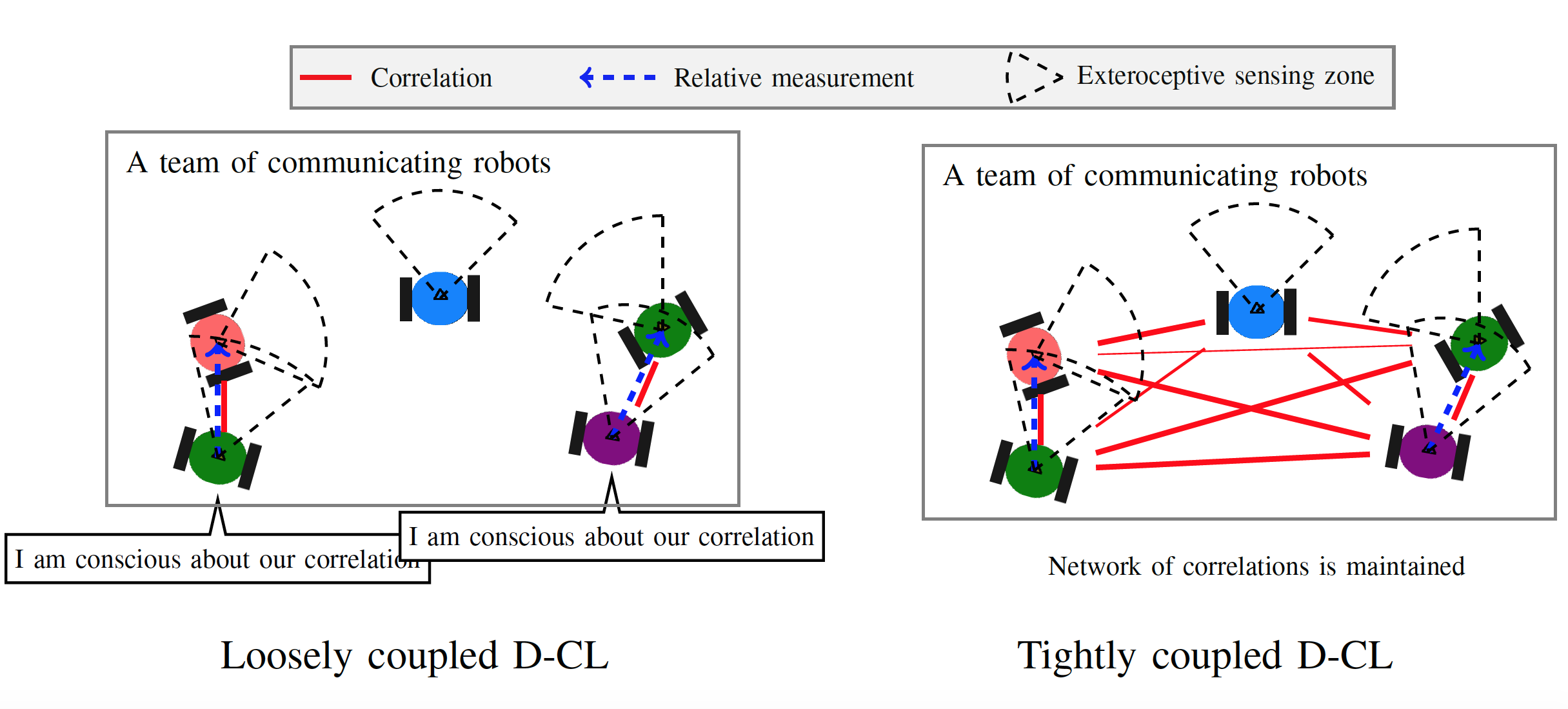}
\end{center}
 \caption{{\small Schematic representation of the D-CL classification
    based on how the past correlations are accounted
    for.}}\label{fig::CLClass}
\end{figure}

Based on the observations that
\begin{itemize}
\item[(a)] past correlations cannot be ignored,
\item[(b)] they are useful to increase the localization accuracy of
  the team,
\item[(c)] the coupling that the correlations create in the state
  estimation of team members is the main challenge in developing a
  decentralized cooperative localization algorithm,
\end{itemize}
one can find, regardless of the technique, two distinct trends in the
design methodology of decentralized cooperative localization
algorithms in the literature.  We term these as ``loosely coupled''
and ``tightly coupled'' decentralized cooperative localization (D-CL)
strategies respectively (see Fig.~\ref{fig::CLClass}).


In the loosely coupled D-CL methodology, only one or both of the
agents involved in a relative measurement update their estimates using
that measurement.  Here, an exact account of the ``network'' of
correlations (see Fig.~\ref{fig::CLClass}) due to the past relative
measurement updates is not accounted for. However, in order to ensure
estimation consistency, some steps are taken to fix this
problem. Examples of loosely coupled D-CL are given
in~\cite{AB-MRW-JJL:09},~\cite{LCC-EDN-JLG-SIR:13},
\cite{POA-CR-RKM:01},~\cite{HL-FN:13} and~\cite{DM-NO-VC:13}.  In the
algorithm of~\cite{AB-MRW-JJL:09}, only the agent obtaining the
relative measurement updates its state. Here, in order to produce
consistent estimates, a bank of extended Kalman filters (EKFs) is
maintained at each agent. Using an accurate book-keeping of the
identity of the agents involved in previous updates and the age of
such information, each of these filters is only updated when its
propagated state is not correlated to the state involved in the
current update equation. Although this technique does not impose a
particular communication graph on the network, the computational
complexity, the large memory demand, and the growing size of
information needed at each update time are its main drawbacks. In the
approach~\cite{LCC-EDN-JLG-SIR:13} it is assumed that the relative
measurements are in the form of relative pose.  This enables the agent
taking the relative measurement to use its current pose estimation and
the current relative pose measurement to obtain and broadcast a pose
and the associated covariance estimation of its landmark agent (the
landmark agent is the agent the relative measurement is taken
from). Then, the landmark agent uses the Covariance Intersection
method (see~\cite{SJJ-JKU:97,SJJ-JKU:01}) to fuse the newly acquired
pose estimation with its own current estimation to increase its
estimation accuracy.  Covariance Intersection for D-CL is also used
in~\cite{POA-CR-RKM:01} for the localization of a group of space
vehicles communicating over a fixed ring topology. Here, each vehicle
propagates a model of the equation of motion of the entire team and,
at the time of relative pose measurements, it fuses its estimation of
the team and of its landmark vehicle via Covariance Intersection.
Another example of the use of split Covariance Intersection is given
in~\cite{HL-FN:13}, for intelligent transportation vehicles localization. Even though the
Covariance Intersection method produces consistent estimations for a
loosely coupled D-CL strategy, this method is known to produce overly
conservative estimates. Another loosely-coupled CL approach is proposed in~\cite{DM-NO-VC:13},
which uses a Common Past-Invariant Ensemble Kalman pose estimation
filter of intelligent vehicles.  This algorithm is very similar to the
decentralized Covariance Intersection data fusion method described
above, with the main difference that it operates with ensembles
instead of with means and covariances. Overall, the loosely coupled algorithms have the advantage of
not imposing any particular connectivity condition on the
team. However, they are conservative by nature, as they do not enable
other agent in the network to fully benefit from measurement updates.

In the tightly coupled D-CL methodology, the goal is to exploit the
``network'' of correlations created across the team (see
Fig.~\ref{fig::CLClass}), so that the benefit of the update can be
extended beyond the agents involved in a given relative
measurement. However, this advantage comes at a potentially higher
computational, storage and/or communication cost.  The dominant trend
in developing decentralized cooperative localization algorithms in
this way is to distribute the computation of components of a
centralized algorithm among team members. Some of the examples for
this class of D-CL is given
in~\cite{NT-SIR-GBG:09,SIR-GAB:02,EDN-SIR-AM:09,KYKL-TDB-HHTL:10,LP-MS-JJL:14}.
In a straightforward fashion, decentralization can be conducted as a
multi-centralized CL, wherein each agent broadcasts its own
information to the entire team. Then, every agent can calculate and
reproduce the centralized pose estimates acting as a fusion
center~\cite{NT-SIR-GBG:09}. Besides a high-processing cost for each
agent, this scheme requires all-to-all agent communication at the time
of each information exchange. A D-CL algorithm distributing
computations of an EKF centralized CL algorithm is proposed
in~\cite{SIR-GAB:02}.  To decentralize the cross-covariance
propagation,~\cite{SIR-GAB:02} uses a singular-value decomposition to
split each cross-covariance term between the corresponding two agents.
Then, each agent propagates its portion.  However, at update times,
the separated parts must be combined, requiring an all-to-all agent
communication in the correction step. Another D-CL algorithm based on
decoupling the propagation stage of an EKF CL using new intermediate
variables is proposed in~\cite{SSK-SR-SM:14-iros}.  But here,
unlike~\cite{SIR-GAB:02}, at update stage, each robot can locally
reproduce the updated pose estimate and covariance of the centralized
EKF after receiving an update message only from the robot that has
made the relative measurement.
Subsequently,~\cite{EDN-SIR-AM:09,LP-MS-JJL:14} present D-CL
strategies using maximum-a-posteriori (MAP) estimation procedure.  In
the former, computations of a centralized MAP is distributed among all
the team members. In the latter, the amount of data required to be
passed between mobile agents in order to obtain the benefits of
cooperative trajectory estimation locally is reduced by letting each
agent to treat the others as moving beacons whose estimate of
positions is only required at communication/measurement times.
The aforementioned techniques all assume that communication messages
are delivered, as prescribed, perfectly all the time. A D-CL approach
equivalent to a centralized CL, when possible, which handles both
limited communication ranges and time-varying communication graphs is
proposed in~\cite{KYKL-TDB-HHTL:10}. This technique uses an
information transfer scheme wherein each agent broadcasts all its
locally available information to every agent within its communication
radius at each time-step. The broadcasted information of each agent
includes the past and present measurements, as well as past
measurements previously~received from other agents. The main drawback
of this method is its high communication and memory cost, which may
not be affordable in applications with limited~communication~bandwidth
and storage resources.

\section{The \ouralg algorithm: a tightly coupled D-CL strategy based
  on Kalman filter decoupling}\label{sec::DCL}

Because of its recursive and simple structure, the EKF is a very
popular estimation strategy. However, as discussed in Section
``Cooperative localization via EKF,''a  naive decentralized
implementation of EKF requires an all-to-all communication at every
time-step of the algorithm. In this section, we describe how by
exploiting a special pattern in the propagation estimation
equations,~\cite{SIR-GAB:02} and~\cite{SSK-SR-SM:14-iros} proposed
tightly coupled\emph{exact} decentralized implementations of EKF for
CL with reduced communication workload per agent. Here, what we mean
by ``exact'' is that if these decentralized implementations are
initialized the same as a centralized EKF, they produce the same state
estimate and the associated state error covariance of the centralized
filter. Our special focus in this section is on the \ouralg
of~\cite{SSK-SR-SM:14-iros}.

The \ouralg algorithm and the algorithm of~\cite{SIR-GAB:02} are
developed based on the observation that, in localization problems, we
are normally only interested in the \emph{explicit} value of the pose
estimate and the error covariance associated with it, while
cross-covariance terms are only required in the update equations.
Such an observation promoted the proposal of the \emph{implicit} tracking
of cross-covariance terms by splitting them into intermediate
variables that can be propagated locally by the agents. Then,
cross-covariance terms can be recovered by putting together these
intermediate variables at any update incidence.  Let the last
measurement update be in time-step $k$ and assume that for $m$
subsequent and consecutive steps no relative measurement incidence
takes place among the team members, i.e., no intermediate measurement
update is conducted in this time interval. In such a scenario, the
propagated cross-covariance terms for these $m$ consecutive steps are
given by
\begin{align}
  \vect{P}_{ij}^{\prpg}(k+l)&=\vect{F}^i(k+l-1)\,\cdots\,
  \vect{F}^i(k)\,\vect{P}_{ij}^{\updt}(k)\,{\vect{F}^j(k)}^\top\,\cdots\,{\vect{F}^j(k+l-1)}^\top,\quad
  l\until{m},\label{eq::propag_central_Expanded-mConseq}
\end{align}
for $i\in\VV$ and $j\in\VV\backslash\{i\}$. That is, at each time step
after $k$, the propagated cross-covariance term is obtained by
recursively multiplying its previous value by the Jacobian of the
system function of agent $i$ on the left and by the transpose of the
Jacobian of the system function of agent $j$ at that time step on the
right. Based on this observation, Roumeliotis and Bekey
in~\cite{SIR-GAB:02} proposed to decompose the last updated
cross-covariance term $\vect{P}_{ij}^{\updt}(k)$ between any agent $i$
and any other agent $j$ of the team into two parts (for example using
the singular value decomposition technique). Then, agent $i$ will be
responsible for propagating the left portion while agent $j$
propagates the right portion. Note that, as long as there is no
relative measurement among team members, each agent can propagate its
portion of the cross-covariance term locally without a need of
communication with others. This was an important result, which led to
a fully decentralized estimation algorithm during the propagation
cycle. However, in the update stage, all the agents needed to
communicate with one and other to put together the split
cross-covariance terms and proceed with the update stage. The approach
to obtain \ouralg, which is outlined below, is also based on the
special pattern that the cross-covariance propagation equations have
in~\eqref{eq::propag_central_Expanded-mConseq}. That is, we also
remove the explicit calculation of the propagated cross-covariance
terms by decomposing them into the intermediate variables that can be
propagated by agents locally. However, this alternative decomposition
allows every agent to update its pose estimate and its associated
covariance in a centralized equivalent manner, using merely an
scalable communication message that is received from the team member
that takes the relative measurement. As such, the \ouralg algorithm
removes the necessity of an all-to-all communication in the update
stage and replaces it with propagating a constant size communication
message that holds the crucial piece of information needed in the
update stage.

In particular, we observe that $\vect{P}_{ij}^{\prpg}(k+l)$
in~\eqref{eq::propag_central_Expanded-mConseq} is composed of the
following 3 parts: (a) $\vect{F}^i(k+l-1)\,\cdots\,\vect{F}^i(k)$ which
is local to agent $i$, (b) the $\vect{P}_{ij}^{\updt}(k)$ that does not
change unless there is relative measurement among the team members,
and (c) ${\vect{F}^j(k)}^\top\,\cdots\,{\vect{F}^j(k+l-1)}^\top$ which
is local to agent~$j$. Motivated by this observation, we propose to
write the propagated
cross-covariances~\eqref{eq::propag_central_Expanded-c} as:
\begin{align}\label{eq::crossCovarAlter}
  \vect{P}_{ij}^{\prpg}(k+1)=\vect{\Phi}^i(k+1)\vect{\Pi}_{ij}(k)
  \vect{\Phi}^j(k+1)^\top,\quad k\in\{0,1,2,\cdots\},
\end{align}
where $\vect{\Phi}^i\in\real^{n^i\times n^i}$, for all $i\in\VV$, is a
time-varying variable that is initialized at
$\vect{\Phi}^i(0)=\vect{I}_{n^i}$ and evolves as:
\begin{equation}\label{eq::phi}
  \vect{\Phi}^i(k+1)=\vect{F}^i(k)\vect{\Phi}^i(k),\quad k\in\{0,1,2,\cdots\},
\end{equation} 
(it is interesting to notice the resemblance of~\eqref{eq::phi} and
the transition matrix for discrete-time systems) and
$\vect{\Pi}_{ij}\in\real^{n^i\times n^j}$, for $i,j\in\VV$ and $i\neq
j$, which is also a time-varying variable that is initialized at
$\vect{\Pi}_{ij}(0)=\vect{0}_{n^i\times n^j}$.  When there is no
relative measurement at time $k+1$,~\eqref{eq::crossCovarAlter}
results into $\vect{\Pi}_{ij}(k+1)=\vect{\Pi}_{ij}(k)$. However, when
there is a relative measurement among the team members
$\vect{\Pi}_{ij}$ must be updated. Next, we derive an expression for
$\vect{\Pi}_{ij}(k+1)$ when there is a relative measurement among team
members at time $k+1$, such that at time $k+2$ one can write
$\vect{P}_{ij}^{\prpg}(k+2)=\vect{\Phi}^i(k+2)\vect{\Pi}_{ij}(k+1)\vect{\Phi}^j(k+2)^\top$.
For this, notice that the update equations~\eqref{eq::S_ab}
and~\eqref{eq::KalmanKi} of the centralized CL algorithm can be
rewritten by replacing the cross-covariance terms
by~\eqref{eq::crossCovarAlter} (recall that in the update stage, we
are assuming that robot $a$ has taken measurement robot robot $b$):
\begin{align}\label{eq::Sab_DCL}
  \vect{S}_{ab}&=\vect{R}^{a}+\Tvect{H}_{a}\vect{P}^{a\prpg}(k+1)\Tvect{H}_{a}^\top+\Tvect{H}_{b}
  \vect{P}^{b\prpg}(k+1)\Tvect{H}_{b}^\top-\nonumber\\
  &\quad\Tvect{H}_{a}\underbrace{\vect{\Phi}^a(k+1)\vect{\Pi}_{ab}(k)\vect{\Phi}^b(k+1)^\top}_{\vect{P}^{\prpg}_{ab}(k+1)}\Tvect{H}_{b}^\top-\Tvect{H}_{b}\underbrace{\vect{\Phi}^b(k+1)\vect{\Pi}_{ba}(k){\vect{\Phi}^a(k+1)}^\top}_{\vect{P}^{\prpg}_{ba}(k+1)}\Tvect{H}_{a}^\top,
\end{align}
and the Kalman gain is
\begin{equation*}
  \vect{K}_i=\vect{\Phi}^i(k+1)\,\vect{\Gamma}_i\,\vect{S}_{ab}\!\!^{-\frac{1}{2}},\quad i\in\VV,
\end{equation*}
where 
\begin{subequations}\label{eq::Di}
\begin{align}
  \vect{\Gamma}_{i}&\!=\!(\vect{\Pi}_{ib}(k){\vect{\Phi}^b}^\top\Tvect{H}_{b}^\top\!-\!\vect{\Pi}_{ia}(k){\vect{\Phi}^a}^\top\Tvect{H}_{a}^\top)\,{\vect{S}_{ab}}\!\!^{-\frac{1}{2}},~i\!\in\!\VV\backslash\{a,\!b\},\label{eq::barD-i}\\
  \vect{\Gamma}_{a}&\!=\!(\vect{\Pi}_{ab}(k){\vect{\Phi}^b}^\top\Tvect{H}_{b}^\top\!-\!(\vect{\Phi}^a)^{-1}\vect{P}^{a\prpg}\Tvect{H}_{a}^\top)\,{\vect{S}_{ab}}\!\!^{-\frac{1}{2}}\!,\label{eq::barD-a}\\
  \vect{\Gamma}_{b}&\!=\!((\vect{\Phi}^{b})^{-1}\vect{P}^{b\prpg}\Tvect{H}_{b}^\top\!-\!\vect{\Pi}_{ba}(k){\vect{\Phi}^a}^\top\Tvect{H}_{a}^\top)\,{\vect{S}_{ab}}\!\!^{-\frac{1}{2}}.\label{eq::barD-b}
\end{align}
\end{subequations}
Generally, $\vect{F}^i(k)$ is invertible for all $k\geq 0$ and
$i\in\VV$. Therefore, $\vect{\Phi}^i(k)$,
for all $k\geq 0$ and $i\in\VV$, is invertible.

Next, for $i\neq j$ and $i,j\in\VV$, we can write the cross-covariance terms~\eqref{eq::RobotCovarUpdate-c} as:
\begin{align*}
\vect{P}_{ij}^{\updt}(k+1)=&~\vect{P}_{ij}^{\prpg}(k+1)\!-\!\vect{K}_i\,\vect{S}_{ab}\,\vect{K}_j^\top\\
=&~\vect{\Phi}^i(k+1)\vect{\Pi}_{ij}(k)\vect{\Phi}^j(k+1)^\top\!\!-\!\big(\vect{\Phi}^i(k+1)\vect{\Gamma}_i\vect{S}_{ab}\!\!^{-\frac{1}{2}}\big)\,\vect{S}_{ab}\,\big(\vect{\Phi}^j(k+1)\vect{\Gamma}_j\vect{S}_{ab}\!\!^{-\frac{1}{2}}\big)^\top\\
=&~~\vect{\Phi}^i(k+1)\big(\vect{\Pi}_{ij}(k)-\vect{\Gamma}_i{\vect{\Gamma}_j}^\top\big)\vect{\Phi}^j(k+1)^\top.
\end{align*}
Let us propose
\begin{align*}
  &\vect{\Pi}_{ij}(k+1)=\vect{\Pi}_{ij}(k)-\vect{\Gamma}_i
  \vect{\Gamma}_j^\top.\end{align*} Then, the cross-covariance
update~\eqref{eq::RobotCovarUpdate-c} can be rewritten as:
\begin{equation}\label{eq::Alter_updateCross}
  \vect{P}_{ij}^{\updt}(k+1)
  =\vect{\Phi}^i(k+1)\,\vect{\Pi}_{ij}(k+1)\,\vect{\Phi}^j(k+1)^\top.
\end{equation}
Therefore, at time $k+2$, the propagated cross-covariances terms for $i\neq j$ and $i,j\in\VV$ are:
\begin{align*}
  \vect{P}_{ij}^{\prpg}(k+2) &=~\vect{F}^i(k+1)\,\vect{P}_{ij}^{\updt}(k+1)\,\vect{F}^j(k+1)^\top\\
  &=~\vect{F}^i(k+1)\vect{\Phi}^i(k+1)\vect{\Pi}_{ij}(k+1)\vect{\Phi}^j(k+1)^\top\vect{F}^j(k+1)^\top\\
  &=~\vect{\Phi}^i(k+2)\,\vect{\Pi}_{ij}(k+1)\,\vect{\Phi}^j(k+2)^\top.
\end{align*}
In short, we can rewrite the propagated and the updated cross-covariance terms of the
centralized EKF CL as, respectively,~\eqref{eq::crossCovarAlter} and~\eqref{eq::Alter_updateCross} for all $k\in\{0,1,\cdots\}$ where 
the variables $\vect{\Phi}^i(k)$'s and
$\vect{\Pi}_{ij}$'s, evolve according to, respectively,~\eqref{eq::phi} and 
\begin{align}\label{eq::bar_Pij}
\vect{\Pi}_{ij}(k+1)=
\begin{cases}
\vect{\Pi}_{ij}(k),&\text{no relative measurement at~} k+1,\\
\vect{\Pi}_{ij}(k)-\vect{\Gamma}_i\,
  \vect{\Gamma}_j^\top,&\text{otherwise},\\
\end{cases}
\end{align}
for  $i,j\in\VV$ and $i\neq j$.

Next, notice that we can write the updated state
estimate and covariance matrix in the new variables as follows, for
$i\in\VV$,
\begin{align}\label{eq::dc_update}
  \Hvect{x}^{i\updt}(k+1)&=\Hvect{x}^{i\prpg}(k+1)+\vect{\Phi}^i(k+1) \,\vect{\Gamma}_i\,\Bvect{r}^{a},\\
  \vect{P}^{i\updt}(k+1)&=\vect{P}^{i\prpg}(k+1)-\vect{\Phi}^{i}(k+1)
  \vect{\Gamma}_{i}\,\vect{\Gamma}_i^\top\vect{\Phi}^i(k+1)^\top\!\!\!,\nonumber
\end{align}
where $\Bvect{r}^{a}=\vect{S}_{ab}\!\!^{-\frac{1}{2}}\vect{r}^{a}$.

Using the alternative
representations~\eqref{eq::crossCovarAlter},~\eqref{eq::Alter_updateCross},
and~\eqref{eq::dc_update} of the EKF CL, the decentralized
implementation~\ouralg is given
in~Algorithm~\ref{alg::ouralg}. 
We develop the \ouralg algorithm by keeping a local copy of
$\vect{\Pi}_{lj}$'s at each agent $i\in\VV$, i.e., $\vect{\Pi}^i_{jl}$
for all $j\in\VV\backslash\{N\}$ and $l\in\{j+1,\cdots,N\}$--because
of the symmetry of the covariance matrix we only need to save, e.g.,
the upper triangular part of this matrix. For example, for a group of
$N=4$ robots, every agent maintains a copy of
$\{\vect{\Pi}^i_{12},\,\vect{\Pi}^i_{13},\,\vect{\Pi}^i_{14},\,\vect{\Pi}^i_{23},\,\vect{\Pi}^i_{24},\,\vect{\Pi}^i_{34}\}$.
During the algorithm implementation, we assume that if
$\vect{\Pi}_{jl}^i$ is not explicitly maintained by agent $i$, the
agent substitutes the value of $(\vect{\Pi}_{lj}^i)^\top$ for it.

\begin{figure}[htbp]
\begin{center}
\includegraphics[trim=0 0 0 0pt,clip,height=1.9
    in]{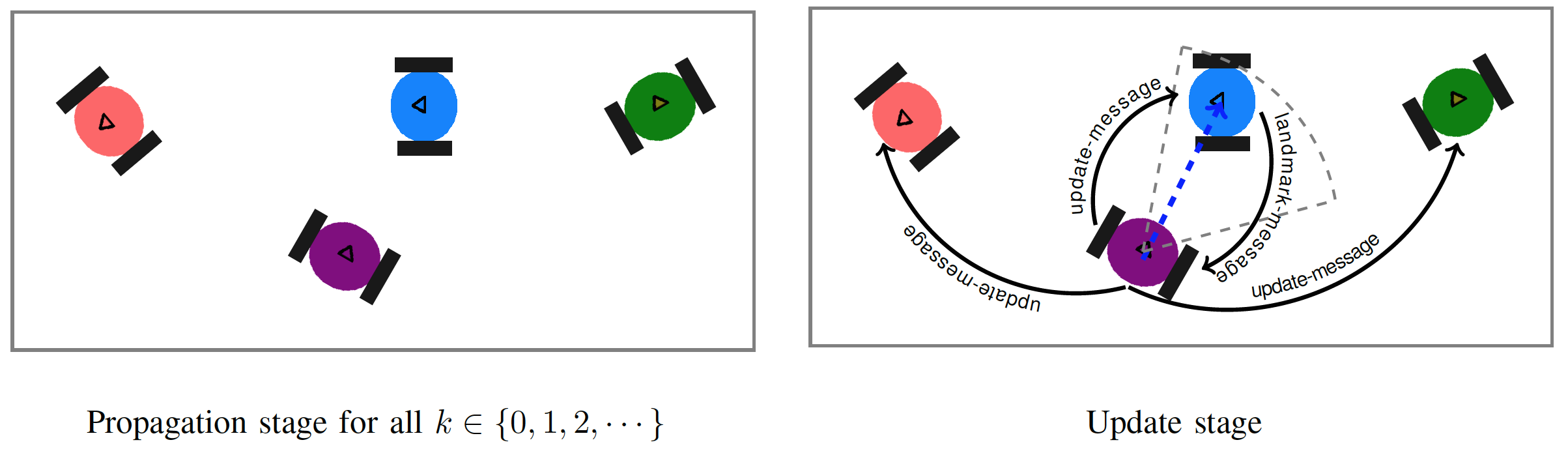}
\end{center}
\caption{{\small The in-network information flow of the \ouralg
    algorithm. In the \ouralg algorithm, communication is only needed in
    the update stage when the team members use a robot-to-robot
    relative measurement feedback to correct their pose
    estimation. Here, we assume that all the team members are in the
    communication range of the interim master
    robot.}}\label{fig::IntrimMaster}
\end{figure}

In \ouralg, every agent $i\in\VV$ initializes its own state estimate
$\Hvect{x}^{i\updt}(0)$, the error covariance matrix
$\vect{P}^{i\updt}(0)$, $\vect{\Phi}^i(0)=\vect{I}_{n^i}$, and its
local copies $\vect{\Pi}^i_{jl}(0)=\vect{0}_{n^j\times n^l}$, for
$j\in\VV\backslash\{N\}$ and $l\in\{j+1,\cdots,N\}$;
see~\eqref{eq::D-CL-init}. At propagation stage, every agent evolves
its local state estimation, error covariance and $\vect{\Phi}^i$,
according to,
respectively,~\eqref{eq::propag_central_Expanded-a},~\eqref{eq::propag_central_Expanded-b},~\eqref{eq::phi};
see~\eqref{eq::D-CL-prpg}. At every time step, when, there is no
exteroceptive measurement in the team, the local updated state
estimates and error covariance matrices are replaced by their
respective propagated counterparts, while $\vect{\Pi}^i_{jl}$'s, to
respect~\eqref{eq::bar_Pij}, are kept unchanged;
see~\eqref{eq::D-CL-updt-nomeas}. When there is a robot-to-robot
measurement,
examining~\eqref{eq::H_ab},~\eqref{eq::reletive_Residual},~\eqref{eq::Sab_DCL},~\eqref{eq::barD-a}
and~\eqref{eq::barD-b} shows that agent $a$, the robot that made the
relative measurement, can calculate these terms using its local
$\vect{\Pi}^i_{jl}$ and acquiring
$\Hvect{x}^{b\prpg}(k\!+\!1)\in\real^{n^b}$,
$\vect{\Phi}^{b}(k\!+\!1)\in\real^{n^b\times n^b}$, and
$\vect{P}^{b\prpg}(k\!+\!1)\in\mathbb{M}_{n^b}$;
see~\eqref{eq::D-CL-landmessag} and~\eqref{eq::D-CL-updt1}.  Then,
agent $a$ can assume the role of the interim master and issue the
update terms for other agents in the team;
see~\eqref{eq::D-CL-updatemessag}. Using this update message and their
local variables, then each agent $i\in\VV$ can
compute~\eqref{eq::barD-i} and use it to obtain its local state
updates of~\eqref{eq::dc_update} and~\eqref{eq::bar_Pij};
see~\eqref{eq::D-CL-updt2}. Figure~\ref{fig::IntrimMaster}
demonstrates the information flow direction between agent while
implementing the \ouralg algorithm.

The inclusion of absolute measurements in the \ouralg is
straightforward. The agent making an absolute measurement is an
interim master that can calculate the \textsl{update-message} using
only its own data and then broadcast it to the team. Next, observe
that the \ouralg algorithm is robust to permanent agent dropouts from
the network. The operation only suffers from a processing cost until
all agents become aware of the dropout. Also, notice that an external
authority, e.g., a search-and-rescue chief, who needs to obtain the
location of any agent, can obtain this location update in any rate
(s)he wishes to by communicating with that agent. This reduces the
communication cost of the~operation.

The \ouralg algorithm works under the assumption that the message from
the agent taking the relative measurement, the interim master, is
reached by the entire team.  Any communication failure results in a
mismatch between the local copies of $\vect{\Pi}_{lj}$ at the agents
receiving and missing the communication message.  The readers are
referred to~\cite{SSK-SR-SM:15} where the authors present a variation
of \ouralg which is robust to intermittent communication message
dropouts.  Such guarantees in~\cite{SSK-SR-SM:15} are provided by
replacing the fully decentralized implementation with a partial
decentralization where a shared memory stores and updates the
$\vect{\Pi}_{lj}$'s.

\subsection{Complexity analysis}\label{sec::Complex_anal}
\vspace{-0.08in} For the sake of an objective performance evaluation,
a study of the computational complexity, the memory usage, as well as
communication cost per agent per time-step of the \ouralg algorithm in
terms of the size of the mobile agent team~$N$ is provided next. At
the propagation state of the \ouralg algorithm, the computations per
agent are independent of the size of the team. However, at the update
stage, for each measurement update, the computation of every agent is
of order $N(N-1)/2$ due to~\eqref{eq::DCL-update-crossco}. As multiple
relative measurements are processed sequentially, the computational
cost per agent at the completion of any update stage depends on the
number of the relative measurements in the team, henceforth denoted by
$N_z$. Then, the computational cost per agent is $O(N_z\times N^2)$,
implying a computational complexity of order $O(N^4)$ for the worst
case where all the agents take relative measurement with respect to
all the other agents in the team, i.e., $N_z=N(N-1)$. The memory cost
per agent is of order $O(N^2)$ which, due to the recursive nature of
the \ouralg algorithm, is independent of $N_z$. This cost is caused by
the initialization~\eqref{eq::D-CL-init} and update
equation~\eqref{eq::DCL-update-crossco}, which are of order
$N(N-1)/2$.

For the analysis of the communication cost, let us consider the case
of a multi-hop communication strategy.  The \ouralg requires
communication only in its update stage, where landmark robots should
broadcast their landmark message to their respective master, and every
agent should re-broadcast any update-message it receives. Let $N_r$ be
the number of the agents that have made a relative measurement at the
current time, i.e., $N_r\leq N$ is the number of current sequential
interim masters. These robots should announce their identity and the
number of their landmark robots to the entire team for sequential update cuing purpose, incurring a communication cost of
order $N_r$ per robots. Next, the team will proceed by sequentially
processing the relative measurements. Every agent can be a landmark of
$N_a\leq N_r$ agents and/or a master of $N_b\leq (N-1)$ agents. The
updating procedure starts by a landmark robot sending its
landmark-message to its active interim master, resulting in a total
communication cost of $O(N_a)$ per landmark robot at the end of update
stage. Every active interim master should pass an update message to
the entire team, resulting in a total communication cost of $O(N_b)$
per robot. Because there are $N_r$ masters, at the end of the update
stage, every robot incurs a communication cost of $O(N_r\times N_b)$
to pass the update messages. Because $N_a,N_r<N_r\times N_b\leq N_z$,
the total communication cost at the end of the update stage is of
order $O(N_z)$ per agent, implying a worst case broadcast cost of
$O(N^2)$ per agent. If the communication range is unbounded, the
broadcast cost per agent is $O(\max\{N_b,N_a\})$, with the worst case
cost of order $O(N)$. The communication message size of each agent in
both single or multiple relative measurements is independent of the
group size $N$. As such for the worst case scenario the
communication message size is of order~$O(1)$.

The results of the analysis above are summarized in
Table~\ref{table::complex} and are compared to those of a trivial
decentralized implementation of the EKF for CL (denoted by T-D-CL) in
which every agent $i\in\VV$ at the propagation stage
computes~\eqref{eq::propag_central_Expanded}--using the broadcasted
$\vect{F}^j(k)$ from every other team member
$j\in\VV\backslash\{i\}$--and at the update stage computes~
\eqref{eq::KalmanKi} and \eqref{eq::RobotCovarUpdate}--using the
broadcast ($a$, $b$, $\vect{r}^a$, $\vect{S}_{ab}$, $\Tvect{H}_a$,
$\Tvect{H}_b$, $\vect{R}^a$, $\vect{P}^{a\prpg}$, $\vect{P}^{b\prpg}$)
from agent $a$ that has made relative measurement from agent
$b$. Agent $a$ calculates $\vect{S}_{ab}$, $\Tvect{H}_a$,
$\Tvect{H}_b$ by requesting ($\Hvect{x}^{b\prpg}$,
$\vect{P}^{b\prpg}$) from agent $b$. We assume that multiple
measurements are processed sequentially and that the communication
procedure is multi-hop. Although the overall cost of the T-D-CL
algorithm is comparable with the \ouralg algorithm, this
implementation has a more stringent communication connectivity
condition, requiring a \emph{strongly connected digraph} topology
(i.e., all the nodes on the communication graph can be reached by
every other node on the graph) at each time-step, regardless of
whether there is a relative measurement incidence in the team. As an
example, notice that the communication graph of
Fig.~\ref{fig::data_propagation} is not strongly connected and as such
the T-D-CL algorithm can not be implemented whereas the \ouralg
algorithm can be. Recall that the \ouralg algorithm needs no
communication at the propagation stage and it only requires an
existence of a spanning tree rooted at the agent making the relative
measurement at the update stage. Finally, the \ouralg algorithm incurs
less computational cost at the propagation stage.

Algorithm~\ref{alg::ouralg-alt} presents an alternative \ouralg
implementation where, instead of storing and evolving
$\vect{\Pi}_{lj}$'s of the entire team, every agent only maintains the
terms corresponding to its own cross-covariances;
see~\eqref{eq::D-Cl-init-alt} and~\eqref{eq::D-Cl-prpg-alt}. For
example in a team of $N=4$, robot $1$ maintains
$\{\vect{\Pi}^1_{12},\vect{\Pi}^1_{13},\vect{\Pi}^1_{14}\}$, robot $2$
maintains $\{\vect{\Pi}^2_{21},\vect{\Pi}^2_{23},\vect{\Pi}^2_{24}\}$,
etc. However, now the interim master $a$ needs to acquire the
$\vect{\Pi}_{bj}$'s from the landmark robot $b$ and calculate and
broadcast $\vect{\Gamma}_i$, $i\in\VV$ to the entire team;
see~\eqref{eq::D-CL-landmessag-alt}, ~\eqref{eq::D-CL-updt1-alt}
and~\eqref{eq::D-CL-updt2-alt}. In this alternative implementation,
the processing and storage cost of every agent is reduced from
$O(N^2)$ to $O(N)$, however the communication message size is
increased from $O(1)$ to $O(N)$.

\subsection{Tightly coupled versus loosely coupled D-CL: a numerical comparison study}
\begin{figure}[htbp]
    \captionsetup[subfigure]{labelformat=empty}
      \unitlength=0.2in 
  \captionsetup[subfloat]{captionskip=-4pt}
\begin{center}
  \subfloat[ {\small Robot 1}] {\includegraphics[trim=0 0 0 0pt,clip,height=1.6
    in]{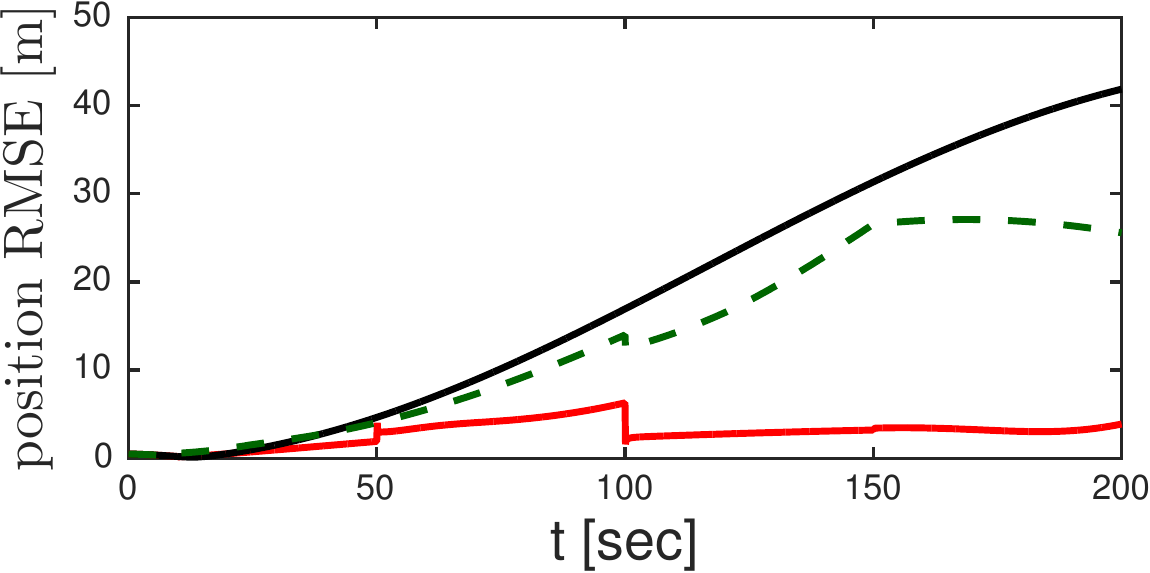}}\\
     \subfloat[{\small Robot 2}]
  {\includegraphics[trim=0 0 0 0pt,clip,height=1.6
    in]{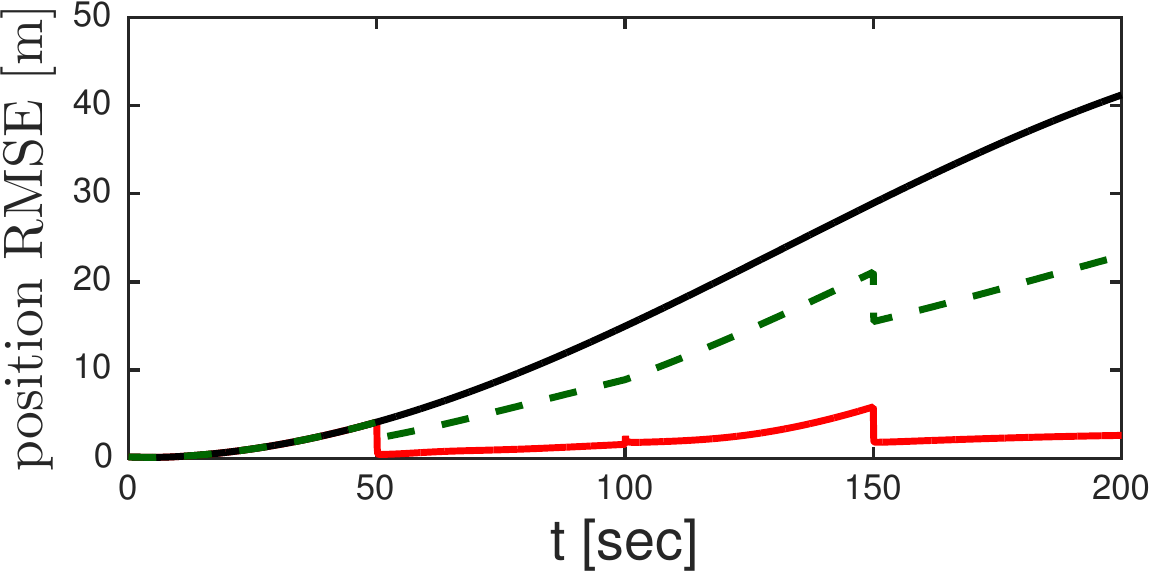}}
    \\
 \subfloat[ {\small Robot3}] {\includegraphics[trim=0 0 0
    0pt,clip,height=1.6
    in]{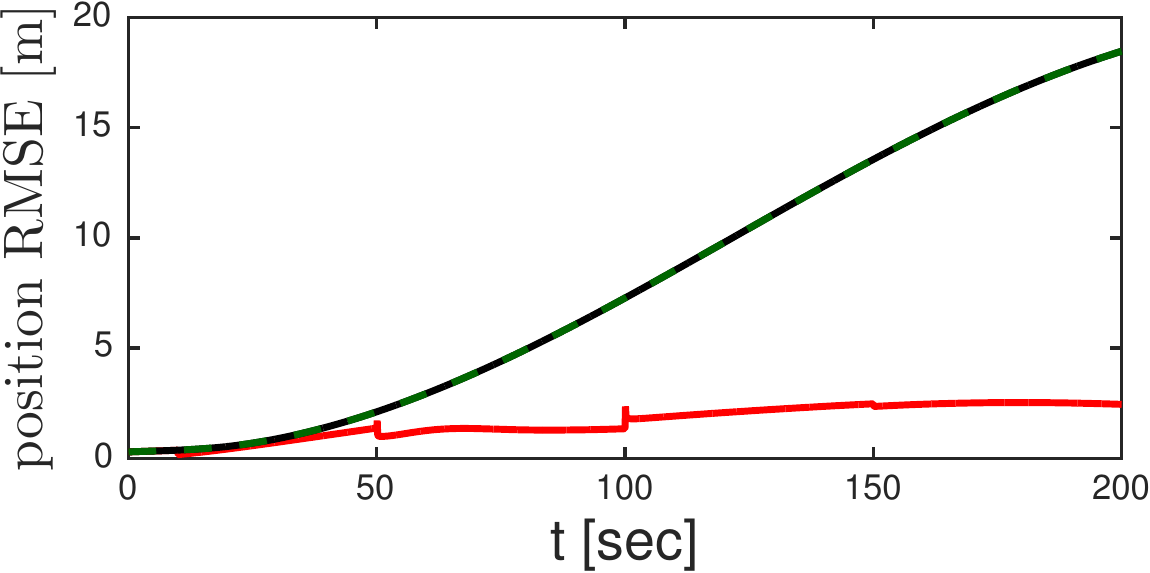}} 
   \end{center}\vspace{-0.4in}
        
   \caption{{\small A comparison study between the positioning
       accuracy of 3 robots employing the \ouralg algorithm (red
       plots), with that from the EKF Covariance-Intersection based CL
       algorithm of~\cite{LCC-EDN-JLG-SIR:13} (dashed green plot). The
       curves in black show the positioning accuracy when the robots
       do not use any CL. As expected, the \ouralg algorithm by keeping an
       accurate account of the cross-covariances produces more
       accurate localization results than the algorithm
       of~\cite{LCC-EDN-JLG-SIR:13} . However, this higher accuracy
       comes with higher communication and processing cost per
       robot. Notice here that using algorithm
       of~\cite{LCC-EDN-JLG-SIR:13} robot $3$ does not get to update
       its estimation equations.}}\label{fig::Fig7}
          \end{figure}

The~\ouralg falls under the tightly coupled D-CL
classification. Fig.~\ref{fig::Fig7} demonstrates the positioning
accuracy (time history of the root mean square error (RMSE) plot for
$50$ Monte Carlo simulation runs) of this algorithm versus the loosely
coupled EKF and Covariance-Intersection based algorithm
of~\cite{LCC-EDN-JLG-SIR:13} in the following scenario.  We consider
the $3$ mobile robots employed in the numerical example of Section
``Cooperative localization via EKF'' with motion as described in that section. For the sensing scenario here,
we assume that, starting at $t = 10$ seconds, robot $3$ takes
persistent relative measurements alternating every $50$ seconds from
robot $1$ to robot $2$ and vice versa. As expected, the tightly coupled
\ouralg algorithm produces more accurate position estimation results
than those of the loosely coupled D-CL algorithm
of~\cite{LCC-EDN-JLG-SIR:13} (similar results can be observed for the
heading estimation accuracy, which is omitted here for brevity). 


In the algorithm of~\cite{LCC-EDN-JLG-SIR:13}, every robot keeps an
EKF estimation of its own pose. When a robot takes a relative pose
measurement from another robot (let us call this robot the interim
master here as well), it acquires the current position estimation and
the corresponding error covariance of the landmark robot. Then, it
uses these along with its own current estimation and the current
relative measurement to extract a new state estimation and the
corresponding error covariance for the landmark robot. After this, the
interim master robot transmits these new estimates to the landmark
robot which uses the Covariance Intersection method to fuse them
consistently to its current pose estimate. It is interesting to notice
that in this particular scenario, even though robot $3$ has been
taking all the relative measurements, it receives no benefit from such
measurements, because only the landmark robots are updating their
estimations. Even though the positioning accuracy of
  algorithm~\cite{LCC-EDN-JLG-SIR:13} is lower, it only requires
  $O(1)$ computational cost per agent as compared to the $O(N^2)$ cost
  of the \ouralg algorithm. However, it also requires more complicated
  calculations to perform Covariance Intersection fusion.  If we
  assume that the communication range of each agent covers the entire
  team, then interestingly the communication cost of these two
  algorithms is the same as both use an $O(1)$ landmark and update
  messages. However, if the communication range is bounded, the
  loosely coupled algorithm of~\cite{LCC-EDN-JLG-SIR:13} offers a more
  flexible and cost effective communication policy.

\vspace{-0.06in}
\section{Conclusions}

\vspace{-0.05in} Here, we presented a brief review on Cooperative
Localization as an strategy to increase the localization accuracy of
team of mobile agents with communication capabilities. This strategy
relies on use of agent-to-agent relative measurements (no reliance on
external features) as a feedback signal to \emph{jointly estimate} the
poses of the team members.  In particular, we discussed challenges
involved in designing decentralized Cooperative Localization
algorithms. Moreover, we presented a decentralized cooperative
localization algorithm that is exactly equivalent to the centralized
EKF algorithm of~\cite{SIR-GAB:02}. In this decentralized algorithm,
the propagation stage is fully decoupled i.e., the propagation is a
local calculation and no intra-network communication is needed. The
communication between agents is only required in the update stage when
one agent makes a relative measurement with respect to another
agent. The algorithm declares the agent made the measurement as
interim master that can, by using the data acquired from the landmark
agent, calculate the update terms for the rest of the team and deliver
it to them by broadcast. Future extensions of this work
  includes concern handling message dropouts and asynchronous
  measurement updates.

\bibliographystyle{unsrt}


\clearpage
\begin{algorithm}[!h]
{\scriptsize
\caption{EKF CL (centralized)}
\label{alg::CEKFCL}
\begin{algorithmic}[1]
\Require
 Initialization ($k=0$): For $i\in\VV$, the algorithm is initialized at
\begin{align*}
& \!\Hvect{x}^{i\updt}\!(0)\!\in\!\real^{n^i}\!\!\!,~\vect{P}^{i\updt}\!(0)\!\in\!\mathbb{M}_{n^i},\vect{P}_{ij}^{\updt}(0)=\vect{0}_{n^i\times n^j},~ j\!\in\!\VV\backslash\{i\}.
\end{align*}

\hspace{-0.38in}\noindent\textbf{Iteration $k$}
\State Propagation: for $i\in\VV$, the propagation equations are:
\begin{subequations}\label{eq::propag_central_Expanded}
\begin{align}
\!\Hvect{x}^{i\prpg}(k\!+\!1)\!&=\vect{f}^i(\Hvect{x}^{i\updt}(k),\vect{u}^i(k)),\label{eq::propag_central_Expanded-a}\\
\!\!\!\!\vect{P}^{i\prpg}(k\!+\!1)\!&=\vect{F}^i(k)\vect{P}^{i\updt}(k)\vect{F}^i(k)\!^\top\!\!\!+\!\vect{G}^i(k)\vect{Q}^i(k)\vect{G}^i(k)\!^\top\!\!\!,\label{eq::propag_central_Expanded-b}\\
\!\!\!\!\!\vect{P}_{ij}^{\prpg}(k\!+\!1)\!&=\vect{F}^i(k)\vect{P}_{ij}^{\updt}(k){\vect{F}^j(k)}\!^\top\!\!,~~ j\in\VV\backslash\{i\}.\label{eq::propag_central_Expanded-c}
\end{align}
\end{subequations}%

\State Update: While there are no relative measurements no update happens, i.e., 
\begin{align*}
\Hvect{x}^{\updt}(k+1)&= \Hvect{x}^{\prpg}(k+1),\quad \vect{P}^{\updt}(k+1)= \vect{P}^{\prpg}(k+1).
\end{align*}
When there is a relative measurement at time-step $k+1$, for example robot $a$ makes a relative measurement of robot $b$, the update proceeds as below. The innovation of the relative measurement 
and its covariance are, respectively, 
\begin{align*}
\vect{r}^{a}&=\vect{z}_{ab}-\vect{h}_{ab}(\Hvect{x}^{a\prpg}(k+1),\Hvect{x}^{b\prpg}(k+1)),
\end{align*} 
and 
\begin{align}\label{eq::S_ab}
\vect{S}_{ab}&=\vect{R}^{a}(k+1)+\Tvect{H}_a(k+1)\vect{P}^{a\prpg}(k+1)\Tvect{H}_a(k+1)^\top+\Tvect{H}_b(k+1) \vect{P}^{b\prpg}(k+1)\Tvect{H}_b(k+1)^\top\nonumber\\ 
&\quad-\Tvect{H}_b(k+1)\vect{P}_{ba}^{\prpg}(k+1){\Tvect{H}_a}(k+1)^\top-\Tvect{H}_a(k+1)\vect{P}_{ab}^{\prpg}(k+1)\Tvect{H}_b(k+1)^\top.
\end{align}
The estimation updates for the centralized EKF are:
\begin{subequations}\label{eq::RobotCovarUpdate}
\begin{align}
\!\Hvect{x}^{i\updt}(k\!+\!1)\!=&\Hvect{x}^{i\prpg}(k\!+\!1)+\vect{K}_i(k\!+\!1)\vect{r}^{a}(k\!+\!1),\label{eq::RobotCovarUpdate-a}\\
\!\vect{P}^{i\updt}(k\!+\!1)\!=&\vect{P}^{i\prpg}\!(k\!+\!1)\!-\!\vect{K}_i(k\!+\!1) \vect{S}_{ab}(k\!+\!1)\vect{K}_i(k\!+\!1)^\top\!\!\!,\label{eq::RobotCovarUpdate-b}\\
\!\vect{P}_{ij}^{\updt}(k\!+\!1)\!=&\vect{P}_{ij}^{\prpg}\!(k\!+\!1)\!-\!\vect{K}_i(k\!+\!1)\vect{S}_{ab}(k\!+\!1)\vect{K}_j(k\!+\!1)^\top\!\!\!,\label{eq::RobotCovarUpdate-c}
\end{align}
\end{subequations}
where $i\in\VV$, $j\in\VV\backslash\{i\}$ and 
\begin{align}\label{eq::KalmanKi}
&\vect{K}_i=(\vect{P}_{i b}^{\prpg}(k+1)\Tvect{H}_b^\top-\vect{P}_{ia}^{\prpg}(k+1)\Tvect{H}_a^\top){\vect{S}_{ab}}^{-1}.
\end{align}

\State $k \leftarrow k+1$
\end{algorithmic}
}
\end{algorithm}

\clearpage

\begin{algorithm}[!t]
{\scriptsize
\caption{\ouralg}
\label{alg::ouralg}
\begin{algorithmic}[1]
\Require
 Initialization ($k=0$): Every agent $i\in\VV$ initializes its filter at
\begin{align}\label{eq::D-CL-init}
& \Hvect{x}^{i\updt}(0)\in\real^{n^i}, \vect{P}^{i\updt}(0)\in\mathbb{M}_{n^i},~~\vect{\Phi}^i(0)=\vect{I}_{n^i},\quad \vect{\Pi}^i_{jl}(0)=\vect{0}_{n^l\times n^j}, ~j\in\VV\backslash\{N\},~ l\in\{j+1,\cdots,N\}.
\end{align}

\hspace{-0.38in}\noindent\textbf{Iteration $k$}
\State Propagation: Every agent $i\in\VV$ propagates the variables below 
\begin{align}\label{eq::D-CL-prpg}
\Hvect{x}^{i\prpg}(k\!+\!1)&\!=\!\vect{f}^i(\Hvect{x}^{i\updt}(k),\vect{u}^i(k)),\quad\vect{P}^{i\prpg}(k\!+\!1)\!=\!\vect{F}^i\!(k)\vect{P}^{i\updt}\!(k)\vect{F}^i\!(k)\!^\top\!\!\!+\!\vect{G}^i\!(k)\vect{Q}^i\!(k)\vect{G}^i\!(k)\!^\top\!\!\!,\quad \vect{\Phi}^i(k\!+\!1)\!=\!\vect{F}^i(k)\vect{\Phi}^i(k).
\end{align}

\State Update: 
while there are no relative measurements in the network, every agent
$i\in\VV$ updates its variables as:
\begin{align}\label{eq::D-CL-updt-nomeas}
\Hvect{x}^{i\updt}(k+1)&=\Hvect{x}^{i\prpg}(k+1),\quad\vect{P}^{i\updt}(k+1)=\vect{P}^{i\prpg}(k+1),\quad \vect{\Pi}^i_{jl}(k+1)=\vect{\Pi}^i_{lj}(k), ~~j\in\VV\backslash\{N\},~ l\in\{j+1,\cdots,N\}.
\end{align}
If there is an agent $a$ that makes a measurement with respect to
another agent $b$, then agent $a$ is declared as the interim master
and acquires the following information from agent $b$: 
\begin{align}\label{eq::D-CL-landmessag}
\textsl{landmark-message}&={\color{blue}\Big(\Hvect{x}^{b\prpg}(k+1), \vect{\Phi}^{b}(k+1), \vect{P}^{b\prpg}(k+1)\Big)}.
\end{align} 
Agent $a$ makes the following calculations upon receiving the
\textsl{landmark-message}:
\begin{subequations}\label{eq::D-CL-updt1}
\begin{align}
\vect{r}^{a}&=\vect{z}_{ab}-\vect{h}_{ab}(\Hvect{x}^{a\prpg},{\color{blue}\Hvect{x}^{b\prpg}}),\\
\vect{S}_{ab}&=\vect{R}^{a}+\Tvect{H}_{a}\vect{P}^{a\prpg}\Tvect{H}_{a}^\top+\Tvect{H}_{b}^\top{\color{blue}\vect{P}^{b\prpg}}\Tvect{H}_{b}-\Tvect{H}_{a}\vect{\Phi}^a\vect{\Pi}^a_{ab}{\color{blue}\vect{\Phi}^b}^\top\Tvect{H}_{b}^\top-\Tvect{H}_{b}{\color{blue}\vect{\Phi}^b}\vect{\Pi}^a_{ba}{\vect{\Phi}^a}^\top\Tvect{H}_{a}^\top,\\
\vect{\Gamma}_{a}&=(({\vect{\Phi}^a})^{-1}{\vect{\Phi}^a}\vect{\Pi}^a_{ab}{\vect{\Phi}^b}^\top\Tvect{H}_{b}^\top-({\vect{\Phi}^a})^{-1}\vect{P}^{a\mbox{-}}\Tvect{H}_{a}^\top){\vect{S}_{ab}}^{-\frac{1}{2}},\quad \vect{\Gamma}_{b}=(({\color{blue}\vect{\Phi}^b})^{-1}{\color{blue}\vect{P^{b\prpg}}}\Tvect{H}_{b}^\top-\vect{\Pi}^a_{ba}{\vect{\Phi}^a}^\top\Tvect{H}_{a}^\top){\vect{S}_{ab}}^{-\frac{1}{2}},
\end{align}
\end{subequations}
where $\Tvect{H}_{a}(k+1)=\Tvect{H}_{a}(\Hvect{x}^{a\prpg},{\color{blue}\Hvect{x}^{b\prpg}})$ and $\Tvect{H}_{b}(k+1)=\Tvect{H}_{b}(\Hvect{x}^{a\prpg},{\color{blue}\Hvect{x}^{b\prpg}})$ are obtained using~\eqref{eq::H_ab}. 

The interim master passes the
following data, either directly or indirectly (by message passing), to
the rest of the agents in the network:
\begin{align}\label{eq::D-CL-updatemessag}
&\textsl{update-message}={\color{red}\Big(a,b,\Bvect{r}^{a}, \vect{\Gamma}_a,\vect{\Gamma}_b,{\vect{\Phi}^b}^\top\Tvect{H}_{b}^\top{\vect{S}_{ab}}^{-\frac{1}{2}},{\vect{\Phi}^a}^\top\Tvect{H}_{a}^\top{\vect{S}_{ab}}^{-\frac{1}{2}}\Big)}.
\end{align}
Every agent $i\in\VV$, upon receiving the \textsl{update-message}, first
calculates, $\forall j\in\VV\backslash\{a,b\}$, using information
obtained at $k$:
\begin{align}\label{eq::D-CL-Di}
\vect{\Gamma}_j&=\vect{\Pi}^i_{jb}{\color{red}{\vect{\Phi}^b}^\top\Tvect{H}_{b}^\top{\vect{S}_{ab}}^{-\frac{1}{2}}}-\vect{\Pi}^i_{ja}{\color{red}{\vect{\Phi}^a}^\top\Tvect{H}_{a}^\top{\vect{S}_{ab}}\!\!^{-\frac{1}{2}}},
\end{align}
and then updates the following variables: 
\begin{subequations}\label{eq::D-CL-updt2}
\begin{align}
\Hvect{x}^{\!i\updt}(k\!+\!1)&=\Hvect{x}^{i\prpg}\!(k\!+\!1)\!+\!\vect{\Phi}^{i}(k\!+\!1) \,\vect{\Gamma}_i\,{\color{red}\Bvect{r}^{a}},
\\ \vect{P}^{\!i\updt}(k\!+\!1)&=\vect{P}^{\!i\prpg}\!(k\!+\!1)\!-\!\vect{\Phi}^{\!i}(k\!+\!1)\vect{\Gamma}_{i}\vect{\Gamma}_i^{\!\top}\vect{\Phi}^i\!(k\!+\!1)^{\!\top},\\
\vect{\Pi}^{\!i}_{jl}(k\!+\!1)&=\vect{\Pi}^{\!i}_{jl}\!(k)\!-\!\vect{\Gamma}_j \vect{\Gamma}_l^\top, \quad j\!\in\!\VV\backslash\{N\}, l\!\in\!\{j+1,\cdots,N\}\label{eq::DCL-update-crossco}.
\end{align}
\end{subequations}

\State $k \leftarrow k+1$
\end{algorithmic}
}
\end{algorithm}

\begin{algorithm}[!t]
{\scriptsize
\caption{Alternative \ouralg (larger communication message size in favor of lower computation and storage cost per agent)}
\label{alg::ouralg-alt}
\begin{algorithmic}[1]
\Require
 Initialization ($k=0$): Every agent $i\in\VV$ initializes its filter at
\begin{align}\label{eq::D-Cl-init-alt}
& \Hvect{x}^{i\updt}(0)\in\real^{n^i}, \vect{P}^{i\updt}(0)\in\mathbb{M}_{n^i},~~\vect{\Phi}^i(0)=\vect{I}_{n^i},\quad \vect{\Pi}^i_{ij}(0)=\vect{0}_{n^i\times n^j}, ~j\in\VV\backslash\{i\}.
\end{align}

\hspace{-0.38in}\noindent\textbf{Iteration $k$}
\State Propagation: Every agent $i\in\VV$ propagates the variables below 
\begin{align}\label{eq::D-Cl-prpg-alt}
\Hvect{x}^{i\prpg}(k\!+\!1)&\!=\!\vect{f}^i(\Hvect{x}^{i\updt}(k),\vect{u}^i(k)),\quad\vect{P}^{i\prpg}(k\!+\!1)\!=\!\vect{F}^i\!(k)\vect{P}^{i\updt}\!(k)\vect{F}^i\!(k)\!^\top\!\!\!+\!\vect{G}^i\!(k)\vect{Q}^i\!(k)\vect{G}^i\!(k)\!^\top\!\!\!,\quad \vect{\Phi}^i(k\!+\!1)\!=\!\vect{F}^i(k)\vect{\Phi}^i(k).
\end{align}

\State Update: 
while there are no relative measurements in the network, every agent
$i\in\VV$ updates its variables as:
\begin{align*}
\Hvect{x}^{i\updt}(k+1)&=\Hvect{x}^{i\prpg}(k+1),\quad\vect{P}^{i\updt}(k+1)=\vect{P}^{i\prpg}(k+1),\quad \vect{\Pi}^i_{ij}(k+1)=\vect{\Pi}^i_{ij}(k), ~~j\in\VV\backslash\{i\}.
\end{align*}
If there is an agent $a$ that makes a measurement with respect to
another agent $b$, then agent $a$ is declared as the interim master
and acquires the following information from agent $b$: 
\begin{align}\label{eq::D-CL-landmessag-alt}
\textsl{landmark-message}&={\color{blue}\Big(\Hvect{x}^{b\prpg}(k+1), \vect{\Phi}^{b}(k+1), \vect{P}^{b\prpg}(k+1), \vect{\Pi}^{b}_{bj}(k) \text{~where~} j\in\VV\backslash\{a,b\}\Big)}.
\end{align} 
Agent $a$ makes the following calculations upon receiving the
\textsl{landmark-message}:
\begin{subequations}\label{eq::D-CL-updt1-alt}
\begin{align}
\vect{r}^{a}&=\vect{z}_{ab}-\vect{h}_{ab}(\Hvect{x}^{a\prpg},{\color{blue}{\Hvect{x}^{b\prpg}}}),\\
\vect{S}_{ab}&=\vect{R}^{a}+\Tvect{H}_{a}\vect{P}^{a\prpg}\Tvect{H}_{a}^\top+\Tvect{H}_{b}^\top{\color{blue}{\vect{P}^{b\prpg}}}\Tvect{H}_{b}-\Tvect{H}_{a}\vect{\Phi}^a\vect{\Pi}^a_{ab}{{\color{blue}\vect{\Phi}^b}}^\top\Tvect{H}_{b}^\top-\Tvect{H}_{b}{{\color{blue}\vect{\Phi}^b}}(\vect{\Pi}^a_{ab})^\top{\vect{\Phi}^a}^\top\Tvect{H}_{a}^\top,\\
\vect{\Gamma}_{a}&=(({\vect{\Phi}^a})^{-1}{\vect{\Phi}^a}\vect{\Pi}^a_{ab}{\color{blue}{\vect{\Phi}^b}}^\top\Tvect{H}_{b}^\top-({\vect{\Phi}^a})^{-1}\vect{P}^{a\mbox{-}}\Tvect{H}_{a}^\top){\vect{S}_{ab}}^{-\frac{1}{2}},\quad \vect{\Gamma}_{b}=(({\color{blue}{\vect{\Phi}^b}})^{-1}{\color{blue}\vect{P^{b\prpg}}}\Tvect{H}_{b}^\top-(\vect{\Pi}^a_{ab})^\top{\vect{\Phi}^a}^\top\Tvect{H}_{a}^\top){\vect{S}_{ab}}^{-\frac{1}{2}},\\
\vect{\Gamma}_j&=({\color{blue}\vect{\Pi}^b_{bj})^\top{\vect{\Phi}^b}^\top}\Tvect{H}_{b}^\top{\vect{S}_{ab}}^{-\frac{1}{2}}-(\vect{\Pi}^i_{aj})^\top{\vect{\Phi}^a}^\top\Tvect{H}_{a}^\top{\vect{S}_{ab}}^{-\frac{1}{2}},\quad  j\in\VV\backslash\{a,b\},
\end{align}
\end{subequations}
where $\Tvect{H}_{a}(k+1)=\Tvect{H}_{a}(\Hvect{x}^{a\prpg},{\color{blue}{\Hvect{x}^{b\prpg}}})$ and $\Tvect{H}_{b}(k+1)=\Tvect{H}_{b}(\Hvect{x}^{a\prpg},{\color{blue}{\Hvect{x}^{b\prpg}}})$ are obtained using~\eqref{eq::H_ab}. 

The interim master passes the
following data, either directly or indirectly (by message passing), to
the rest of the agents in the network:
\begin{align}\label{eq::D-CL-updatemessag-alt}
&\textsl{update-message}={\color{red}\Big(a,b,\Bvect{r}^{a},\vect{\Gamma}_1,\cdots,\vect{\Gamma}_N\Big)}.
\end{align}
Every agent $i\in\VV$, upon receiving the \textsl{update-message}, updates the following variables: 
\begin{subequations}\label{eq::D-CL-updt2-alt}
\begin{align}
\Hvect{x}^{\!i\updt}(k\!+\!1)&=\Hvect{x}^{i\prpg}\!(k\!+\!1)\!+\!\vect{\Phi}^{i}(k\!+\!1) \,{\color{red}\vect{\Gamma}_i\,\Bvect{r}^{a}},
\\ \vect{P}^{\!i\updt}(k\!+\!1)&=\vect{P}^{\!i\prpg}\!(k\!+\!1)\!-\!\vect{\Phi}^{\!i}(k\!+\!1){\color{red}\vect{\Gamma}_{i}\vect{\Gamma}_i^{\!\top}}\vect{\Phi}^i\!(k\!+\!1)^{\!\top},\\
\vect{\Pi}^{\!i}_{ij}(k\!+\!1)&=\vect{\Pi}^{\!i}_{ij}\!(k)\!-\!{\color{red}\vect{\Gamma}_i \vect{\Gamma}_j^\top},\quad j\in\VV\backslash\{i\}. \label{eq::DCL-update-crossco}
\end{align}
\end{subequations}

\State $k \leftarrow k+1$
\end{algorithmic}
}
\end{algorithm}

\clearpage
\setlength\extrarowheight{2pt}
{\scriptsize
 \begin{table}[h]
  \caption{{\footnotesize Complexity analysis per agent of the \ouralg algorithm (denoted by IM-D-CL) compared to that of the trivial decentralized implementation of EKF for CL (denoted by T-D-CL) introduced in Subsection~\ref{sec::Complex_anal}.}
  }\label{table::complex}\vspace{-0.06in}
  \centering
  \tabcolsep=0.05cm
    \begin{tabular}{| m{2.1cm} || c | c ||c|c||c|c||c|c||p{2cm}|p{1.6cm}|}
    \hline  
      ~& \multicolumn{2}{ |c| }{Computation}&  \multicolumn{2}{ |c| }{Storage} & \multicolumn{2}{ |c| }{Broadcast$^{\star}$}&  \multicolumn{2}{ |c| }{Message Size} & \multicolumn{2}{ |c| }{Connectivity}\\ \hline
    Algorithm&                                             IM-D-CL                        &T-D-CL                        &IM-D-CL     &T-D-CL           & IM-D-CL                             &T-D-CL        &IM-D-CL      &T-D-CL &~IM-D-CL~      &~~T-D-CL  \\ \hline
    Propagation&                                        $O(1)$                      &$O(N^2)$                     & $O(N^2)$  & $O(N^2)$      & $0$                                    &$O(N)$         &  $0$          &$O(1)$ & None&\multirow{3}{*}{\begin{minipage}{1.8cm}\vspace{5pt}strongly connected digraph\end{minipage}}  \\ \cline{1-10} 
    Update per $N_z$ relative measur.\!&    $O(N_z\!\times \!N^2)$     & $O(N_z\!\times\! N^2)$   & $O(N^2)$ &  $O(N^2)$      &  $O(N_z)$ & $O(N_z)$    &  $O(1)$    & $O(1)$& \multirow{2}{*}{\begin{minipage}{1.5cm}\vspace{2pt}interim master can reach all the agents
    \end{minipage}}&  ~ \\ \cline{1-9}
    Overall worst case&                              $O(N^4)$                      & $O(N^4)$                    & $O(N^2)$ &  $O(N^2)$      &  $O(N^2)$       &$O(N^2)$ & $O(1)$        & $O(1)$ &~&~ \\ \hline
    \end{tabular}
    \noindent$^*$Broadcast cost is for multi-hop communication. If the communication range is unbounded, the broadcast cost per agent is $O(\max\{N_b,N_a\})$ with the worst cost~of~$O(N)$.\vspace{-0.18in}
\end{table}}

\clearpage

\clearpage

\clearpage

\clearpage

\clearpage

\clearpage

\clearpage
\begin{center}
\textbf{Sidebar 1}

\textbf{Further Reading} 
\end{center}
A performance analysis of an EKF CL for a team of homogeneous robots
moving on a flat terrain, with the same level of uncertainty in their
proprioceptive measurements and exteroceptive sensors that measure
relative pose, is provided in~\cite{SIR-AIM:04-s} and
~\cite{AIM-SIR:06-s}.  Interestingly,~\cite{SIR-AIM:04-s} shows that
the rate of uncertainty growth decreases as the size of the robot team
increases, but is subject to the law of diminishing returns.
Moreover,~\cite{AIM-SIR:06-s} shows that the upper bound on the rate
of uncertainty growth is independent of the accuracy or the frequency
of the robot-to-robot measurements.  The consistency of EKF CL from
the perspective of observability is studies
in~\cite{GH-NT-AIM-SIR:11-s}. Huang et
al.~in~\cite{GH-NT-AIM-SIR:11-s} analytically show that the
error-state system model employed in the standard EKF CL always has an
observable subspace of higher dimension than that of the actual
nonlinear CL system. This results in an unjustified reduction of the
EKF covariance estimates in directions of the state space where no
information is available, and thus leads to inconsistency.  To address
this problem, Huang et al.~in~\cite{GH-NT-AIM-SIR:11-s} adopt an
observability-based methodology for designing consistent estimators in
which the linearization points are selected to ensure a linearized
system model with an observable subspace of the correct dimension.
More results on observability analysis of CL can be found
in~\cite{SIR-GAB:02, AM-RS:05-s, RS-RWB-CNT-SQ:12-s}.  The use of an
observability analysis to explicitly design an active local path
planning algorithm for unmanned aerial vehicles implementing a
bearing-only CL is discussed in~\cite{RS:Thesis-s}.  The necessity for
an initialization procedure for CL is discussed
in~\cite{NT:Thesis-s}. There it is shown that, because of system
nonlinearities and the periodicity of the orientation, initialization
errors can lead to erroneous results in covariance-based filters.  An
initialization procedure for the state estimation in a CL scenario
based on ranging and dead reckoning is studied in~\cite{JON-PH:13-s}.

\clearpage

\subsection{Authors Information} 

\emph{\textbf{Solmaz S. Kia}} is an Assistant Professor in the Department of Mechanical and Aerospace Engineering, University of California, Irvine (UCI). She obtained her Ph.D. degree in Mechanical and Aerospace Engineering from UCI, in 2009, and her M.Sc. and B.Sc. in Aerospace Engineering from the Sharif University of Technology, Iran, in 2004 and 2001, respectively. She was a senior research engineer at SySense Inc., El Segundo, CA from Jun. 2009-Sep. 2010. She held postdoctoral positions in the Department of Mechanical and Aerospace Engineering at the UC San Diego and UCI. Dr. Kia's main research interests, in a broad sense, include distributed optimization/coordination/estimation, nonlinear control theory and probabilistic robotics.

\emph{\textbf{Stephen Rounds}} is a Research Engineer with NavCom Technology, a John Deere company. He is responsible for identifying and supporting new navigation technologies for the company and developing and maintaining the intellectual property portfolio of the company. Prior to working with John Deere, Mr. Rounds worked with multiple defense contractors, with a special emphasis on GPS-denied navigation, GPS anti-jamming protection, and other sensor fusion applications. Mr. Rounds holds a B.S. degree in physics from Stevens Institute of Technology in Hoboken, N.J., and an M.S. degree in nuclear physics from Yale University.  He is the Chairman of the Southern California section of the Institute Of Navigation, holds multiple patents in the navigation field, and has numerous publications in various aerospace and navigation forums.

\emph{\textbf{Sonia Mart{\'\i}nez}} is a Professor with the department of Mechanical and
Aerospace Engineering at the University of California, San
Diego. Dr. Martinez received her Ph.D. degree in Engineering
Mathematics from the Universidad Carlos III de Madrid, Spain, in May
2002. Following a year as a Visiting Assistant Professor of Applied
Mathematics at the Technical University of Catalonia, Spain, she
obtained a Postdoctoral Fulbright Fellowship and held appointments at
the Coordinated Science Laboratory of the University of Illinois,
Urbana-Champaign during 2004, and at the Center for Control, Dynamical
systems and Computation (CCDC) of the University of California, Santa
Barbara during 2005.  In a broad sense, Dr. Mart{\'\i}nez's main research
interests include the control of networked systems, multi-agent
systems, nonlinear control theory, and robotics.  For her work on the
control of underactuated mechanical systems she received the Best
Student Paper award at the 2002 IEEE Conference on Decision and
Control. She was the recipient of a NSF CAREER Award in 2007. For the
paper ``Motion coordination with Distributed Information," co-authored
with Jorge Cort\'es and Francesco Bullo, she received the 2008 Control
Systems Magazine Outstanding Paper Award. She has served on the
editorial boards of the European Journal of Control (2011-2013), and
currently serves on the editorial board of the Journal of Geometric
Mechanics and IEEE Transactions on Control of Networked Systems.

\end{document}